\documentclass[11pt]{article}
\usepackage{packages}
\usepackage{statistics-macros}
\usepackage{editing-macros}
\usepackage{formatting}
\usepackage{caption}
\usepackage{subcaption}

\title{Pretraining Data Mixtures Enable Narrow Model Selection Capabilities in Transformer Models}
\author{Steve Yadlowsky, Lyric Doshi, Nilesh Tripuraneni\\\texttt{\{yadlowsky, lyric, nileshtrip\}@google.com}\\Google DeepMind}
\date{\today}

\begin{document}

\maketitle

\maketitle

\begin{abstract}
Transformer models, notably large language models (LLMs), have the remarkable ability to perform in-context learning (ICL) -- to perform new tasks when prompted with unseen input-output examples without any explicit model training. In this work, we study how effectively transformers can bridge between their pretraining data mixture, comprised of multiple distinct task families, to identify and learn new tasks in-context which are both inside and outside the pretraining distribution. Building on previous work, we investigate this question in a controlled setting, where we study transformer models trained on sequences of $(x, f(x))$ pairs rather than natural language. Our empirical results show transformers demonstrate near-optimal unsupervised model selection capabilities, in their ability to first in-context identify different task families and in-context learn within them when the task families are well-represented in their pretraining data. However when presented with tasks or functions which are out-of-domain of their pretraining data, we demonstrate various failure modes of transformers and degradation of their generalization for even simple extrapolation tasks. Together our results highlight that the impressive ICL abilities of high-capacity sequence models may be more closely tied to the coverage of their pretraining data mixtures than inductive biases that create fundamental generalization capabilities.
\end{abstract}

\section{Introduction}
One of the impressive capabilities demonstrated by large language models is their ability to do few-shot learning by providing examples \emph{in-context} and asking the model to generate a response to follow the final input provided \citep{NEURIPS2020_1457c0d6}. Researchers have taken the underlying machine learning technology, transformer models, and demonstrated that they can perform also in-context learning tasks in domains other than language \citep{garg2022can,akyurek2022learning,li2023transformers}. In these works, they demonstrate the ability of transformers to learn high-dimensional and non-linear functions of the inputs from in-context examples, often matching or exceeding the performance of state-of-the-art machine learning models tuned for the purpose of learning those functions. For example, \citet{garg2022can} showed that after pretraining on sparse linear data, a transformer network can in-context learn unseen sparse linear functions as well as the Lasso, which is known to be statistically optimal for data being modeled.

In these models, as in the large language models that they are designed to reflect, pretraining (or fine-tuning) the model with relevant data to teach the model how to perform in-context learning is critical to enabling this capability. In this work, we focus in on a specific aspect of this pretraining process---the data used in pretraining---and investigate how it affects the few-shot learning capabilities of the resulting transformer model.

The few-shot learning setup that we study follows \citet{garg2022can}, where the goal is to use a set of provided inputs and labels, $((\x_1, \outs{1}),\allowbreak (\x_2, \outs{2}),\allowbreak \dots (\x_n, \outs{n}))$ to make a prediction about the label $\outs{n+1}$ for a new input $\x_{n+1}$. The number of examples provided is small relative to the amount of data used to pretrain the meta-learning model used to perform this task (hence the ``few-shot'' nomenclature). A common approach to apply sequence models for few-shot learning is first to pass the examples in sequentially, alternating inputs and labels, as $(\x_1, \outs{1}, \x_2, \outs{2}, \dots, \x_n, \outs{n+1})$. Finally, the test input point $\x_{n+1}$ is passed as the final element of the sequence, and model prediction for the next item in the sequence is treated as the predicted label. Previous work \citep{garg2022can, akyurek2022learning, li2023transformers} shows that transformer models are capable of learning many types of data distributions for $(\x, \outs{{}})$ pairs and investigate transformer models' ability to make such predictions.

Training the model to be capable of such predictions requires fitting the model on many sequences of the form $\s_i = (\x_{1,i}, \outs{1,i}, \x_{2,i}, \dots, \x_{n,i}, \outs{n,i}, \x_{n+1,i}, \outs{n+1, i})$. Each example in the sequence is drawn using the same function $f$, %
and each sequence uses a different function $f$ drawn from some distribution $\mathcal{D(F)}$ over function class $\mathcal{F}$. 
We investigate interactions between pretraining data composition and transformers' abilities to few-shot learn related tasks. 

Our contributions are as follows:
\begin{itemize}
    \item We pretrain transformer models for in-context learning using a mixture of multiple distinct function classes and characterize the model selection behavior exhibited.
    \item We study the in-context learning behavior of the pretrained transformer model on functions that are ``out-of-distribution'' from the function classes in the pretraining data. 
    \item In the regimes studied, we find strong evidence that the model can perform model selection among pretrained function classes during in-context learning at little extra statistical cost, but limited evidence that the models' in-context learning behavior is capable of generalizing beyond their pretraining data.
\end{itemize}

\section{Preliminaries}

Transformers are sequence models that provide next-token predictions conditional on the previous sequence tokens. We consider a data-generating model where $d$-dimensional covariates are drawn $\x_i \sim \cN(0, \I_d)$ and a  (random) function $f \sim \cDF$ is sampled from a distribution over function classes. Like \citet{garg2022can} and \citet{akyurek2022learning}, we frame the ICL problem as providing a single prompt sequence\footnote{ We assume that the inputs and outputs are all represented as real scalars or vectors. If not in the same dimension, we can project the lower dimensional items into the higher dimensional space, filling in zeros for the empty dimensions.} $s = (\x_1, f(\x_1), \x_2, f(\x_2), \hdots \x_n, f(\x_n), \x_{n+1})$ to the model (i.e. a transformer) and generating a prediction for $f(\x_{n+1})$: $\tilde{f}(\x_{n+1})$. We refer to the problem of predicting the next token as in-context learning. The performance of an in-context learner is judged by its predictive squared-loss $\mathbb{E}[(\tilde{f}(\x_{n+1})-f(\x_{n+1}))^2]$, with the expectation taken over the randomness in the prompt and query.

\citet{garg2022can} demonstrated that by pretraining a transformer model on simulated prompts represented as such sequences, the model is able to in-context learn unseen functions drawn from the same function class at test-time. For example, they demonstrate such pretrained transformers are able to perform as well as state-of-the-art ML methods on linear function classes (with both sparse and dense coefficient vectors) when pretrained on data generated from linear functions, decision trees when pretrained on data generated from decisions trees, and ReLU networks when pretrained on data generated from ReLU networks. \citet{akyurek2022learning} study transformers' ability to learn linear models in-context and provide mechanistic interpretations of how they may be performing such learning.  \citet{li2023transformers} studies generalization properties of transformers in this setting and demonstrates similar results for linear dynamical systems. \citet{raventos2023pretraining} investigates the role of pretraining function diversity for in-context learning in the setting of pure linear regression -- arguing that a sufficiently diverse distribution over linear tasks in pretraining is needed for ICL at test-time. Closest to our work on model selection is that of \citet{bai2023transformers}, which explores transformers' abilities to perform model selection on empirical grounds and provides rigorous theoretical guarantees for transformers generalization properties in pretraining and their downstream in-context prediction performance. The guarantees and empirics in \citet{bai2023transformers} are restricted to explorations on model selection amongst different linear function class families -- namely studying model selection across evenly weighted task mixtures of linear regression with different label noise strengths and linear/logistic regression.

In our setting, given a data source $\datasrc{}$ containing sequences $\s_i = (\x_{i,1},\allowbreak \outs{i,1},\allowbreak ...,\allowbreak \x_{i,n},\allowbreak \outs{i,n})$, we pretrain the parameters $\theta$ of the transformer model $m_\theta(\s)$ by performing loss minimization on the ``teacher forcing'' objective
\begin{equation}
    \label{eq:loss}
    \mathcal{L}(\theta) = \frac{1}{|\datasrc{}|} \sum_{\s_i \in \datasrc{}} \sum_{j = 1}^n \ell\left(\outs{i, j}, m_{\theta}(\s_{i, 1:j})\right),
\end{equation}
for the squared-loss $\ell$, where we use $\s_{i, 1:j}$ to refer to the $i$-th sequence up to (but not including) the $j$-th output $\outs{i,j}$. We use a 9.5M parameter decoder model with 12 layers, 8 attention heads, and a 256-dimensional embedding space as in \citet{garg2022can}. Details on the model and our training setup are in Appendix ~\ref{app:architecture}.

The focus of this paper is understanding how the construction of the data source $\datasrc{}$ affects the in-context learning abilities of the model in the controlled setting of learning function classes. In this case of studying a single function class family as in \citet{garg2022can, akyurek2022learning} and \citet{li2023transformers}, for a linear data-generating model, a single $f$ can be effectively sampled by drawing $\bbeta \sim \mathcal{N}(0, \I_d)$ and defining $f(\x) = \bbeta^\top \x$ for use in a sequence $\s_i$.

In this work, we use data mixtures which combine together examples generated from multiple distinct function families.  We consider several distributions over base function classes $\cDF$: (a) $\cDdense$: dense linear functions, (b) $\cDsparse{nnz}$: sparse linear functions with \emph{nnz} non-zero coordinates, (c) $\cDrelu$: two-layer ReLU networks, and (d) $\cDsine$: sinusoidal functions. Additional details on how these base distributions over function classes are generated are provided in the Appendix ~\ref{app:data_gen}. Mixture distributions over function classes take the form: $\cDF = w \cdot \cDFa + (1-w) \cdot \cDFb$ for selected function classes $\cFa$ and $\cFb$. Each training prompt sequence is constructed by randomly selecting a function class from the mixture based on $w$ and sampling from that base class. For example, for $w=.25$ with selected function classes $\cDdense$ and $\cDsparse{2}$, a sequence $\s_i$ is drawn from $\cDdense$ with probability $0.25$ and from $\cDsparse{2}$ with probability 0.75.

\citet{garg2022can} argues transformers generalize well to tasks/function drawn from the same distribution as the training data. However, one general open question is how these models perform on examples that are out-of-distribution from the training data. In the setting studied here, we interpret the generalization question as the following: \emph{``Can a model generate good predictions with in-context examples from a function not in any of the base function classes seen in the pretraining data mixture?"} We build to this question by first studying the abilities of transformers to perform model selection between different function class families seen in pretraining in Section~\ref{sec:model_selection}. We then transition to answering the OOD-generalization question for a few important cases in Section~\ref{sec:sine}.

\section{Model Selection Phenomena}
\label{sec:model_selection}

One question that comes up when pretraining data mixture of different function classes is \emph{``how does the model select between different function classes when presented with in-context examples in the support of the pretraining mixture?''} We address this question here from an empirical point of view.\footnote{The mechanistic question about how these empirical phenomena come to be is interesting, but we do not address it here. We believe that first clearly documenting the current phenomena is important for the research community.} In this section, we find that the models make optimal (or nearly so) predictions after seeing in-context examples from a function class which is a member of the pretraining data mixture. We also observe how models perform on functions that are not cleanly part of any single component function class before exploring some functions that are definitively out-of-distribution from all pretraining data in Section~\ref{sec:sine}.

We begin with the study of linear functions, which have received a significant attention in this area of in-context learning. \citet{garg2022can} show that transformers pretrained on linear functions perform nearly optimally at in-context learning on new linear functions. In particular, they consider two models: one trained on dense linear functions (where all of the coefficients of the linear model are non-zero), and one trained on sparse linear functions (where say only 2 of the 20 coefficients are non-zero). Each model performs correspondingly as well as linear regression and Lasso regression on new dense and sparse linear functions, respectively. We additionally compare these two models to a model pretrained on a mixture of both sparse and dense linear functions.

\begin{figure}[thb]
\centering     %
\includegraphics[width=0.6\linewidth]{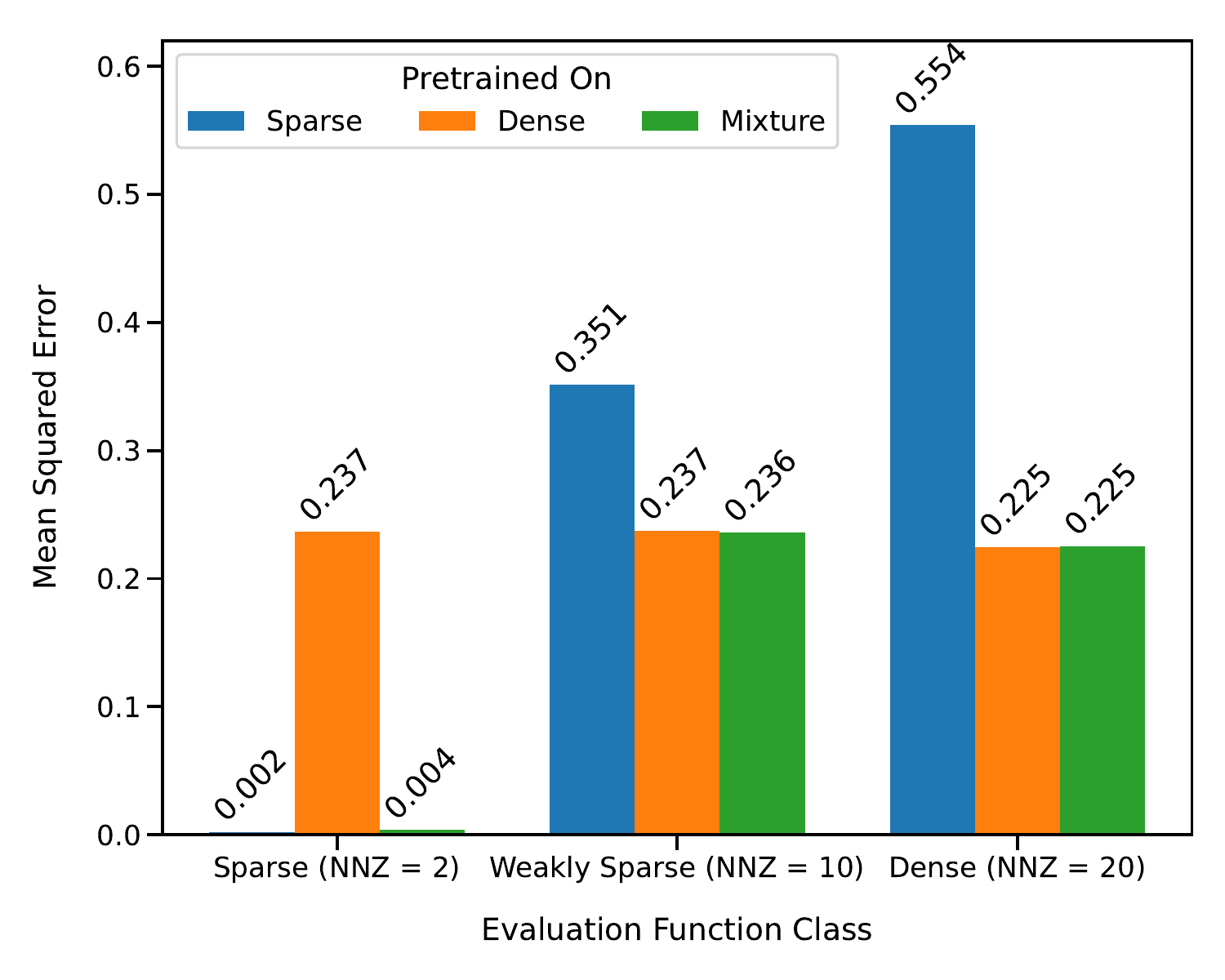}
\caption{Comparison of models with three different pretraining data compositions (denoted in bar colors), on three different evaluation functions passed in-context (denoted in x-axis groupings). In this case mixture distribution in pretraining is evenly weighted with $w=.5$ in $\cDF = w \cdot \cDdense + (1-w) \cdot \cDsparse{2}$. 16 in-context examples provided to all models, with a input data dimension of 20, so that exact recovery of dense coefficients is not possible.}
\label{fig:linear_linear_compare}
\end{figure}

\begin{figure}[!htb]
\centering
\caption{\label{fig:dense_sparse_sparse_learning_curves}ICL learning curves  displaying mean-squared error (MSE) for evaluations on prompts drawn from $\cDdense$ (left) and $\cDsparse{2}$ (right). Transformers were pretrained on mixtures of $w \cdot \cDdense + (1-w) \cdot \cDsparse{2}$. Different curves correspond to different $w$.}
\begin{subfigure}{.45\textwidth}
    \includegraphics[scale=0.3]{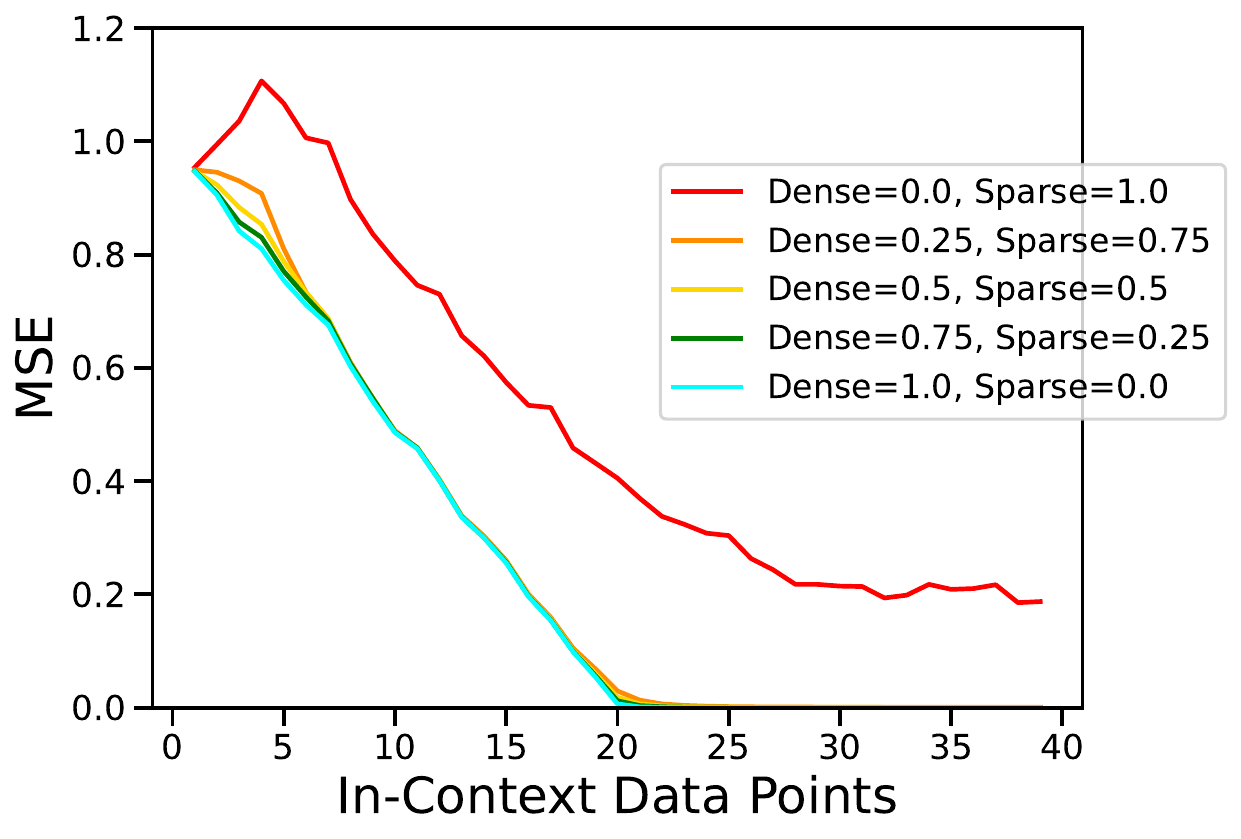}
    \caption{Evaluated on prompts from $\cDdense$}
    \label{subfig:dense_sparse_sparse_no_instr}
\end{subfigure}
\begin{subfigure}{.45\textwidth}
    \includegraphics[scale=0.3]{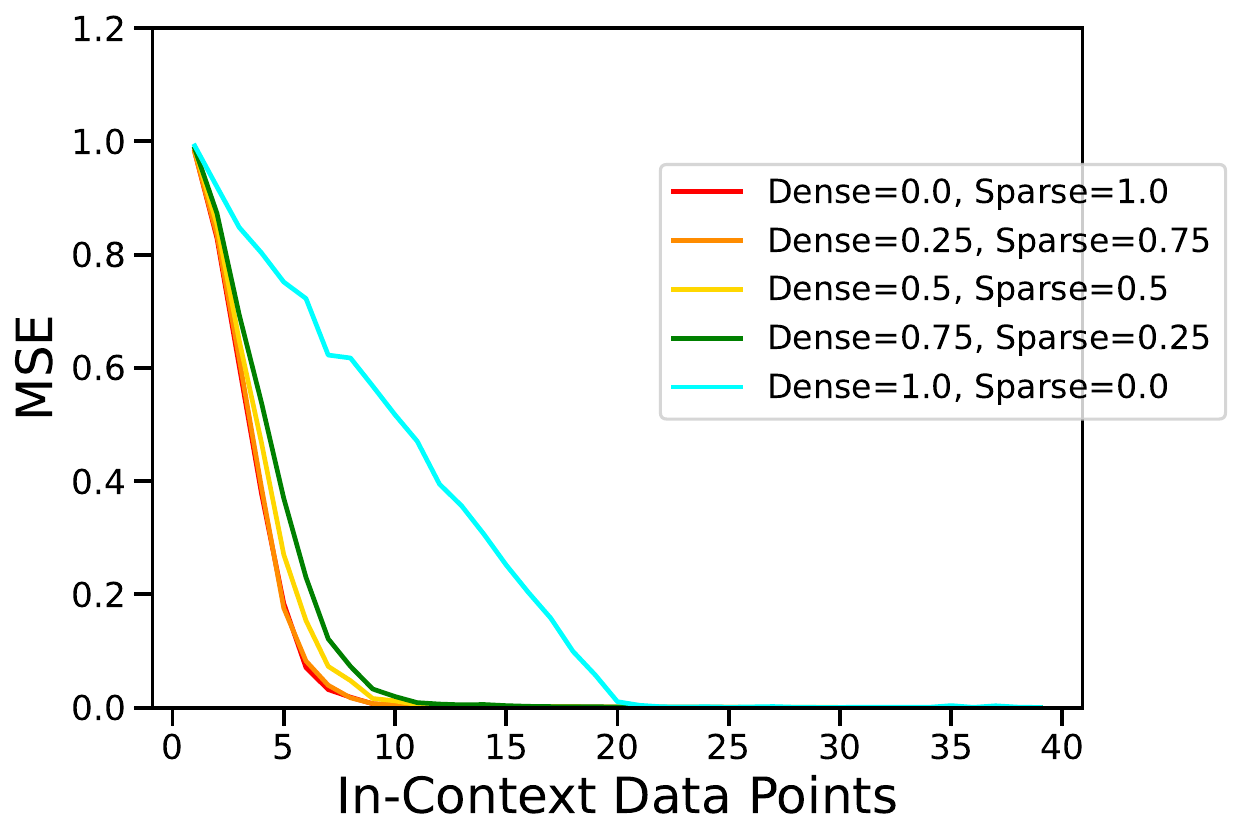}
    \caption{Evaluated on prompts from $\cDsparse{2}$.}
    \label{subfig:dense_sparse_sparse_instr}
\end{subfigure}
\end{figure}

Figure~\ref{fig:linear_linear_compare} shows that the model pretrained on a $\cDF = 0.5 \cdot \cDdense + 0.5 \cdot \cDsparse{2}$ mixture performs similarly at in-context learning as models pretrained on only one function class. Since the model pretrained on the mixture performs similarly to the models shown by \citet{garg2022can} to be theoretically optimal, we infer that this model is nearly optimal, as well. The ICL learning curves in Figure~\ref{fig:dense_sparse_sparse_learning_curves} show us that this in-context model selection ability is relatively uniform with respect to the number of in-context examples provided. In Figure~\ref{fig:dense_sparse_sparse_learning_curves}, we also see that ICL learning curves for pretraining data mixtures with various non-trivial weight $w$ (or $1-w$) for a given function class nearly match the optimal baseline sample complexity compared to pretraining a model purely on that function class. We observe only small deviations and these decay quickly as ICL sample count increases, matching the behavior in Figure~\ref{fig:linear_linear_compare} which corresponds to a single point on the ICL learning curve.
Figure~\ref{fig:dense_sparse_sparse_learning_curves} also demonstrates that transformer model ICL generalization suffers out-of-distribution. Even though dense and sparse linear classes are both linear functions, we can see the poor performance of the red curve in Figure~\ref{subfig:dense_sparse_sparse_no_instr} (which corresponds to a transformer pretrained on only sparse linear functions and evaluated on dense linear data) and vice-versa for the teal curve in Figure~\ref{subfig:dense_sparse_sparse_instr}. We see similar behavior for other nonlinear function classes, as detailed in Appendix~\ref{app:model_selection}. We additionally briefly explore the effect of model size on model selection in the later part of Appendix~\ref{app:model_selection}.

\begin{figure}[thb]
\centering     %
\caption{Comparison of models with three different pretraining data compositions across different levels of sparsity in the linear function evaluated (i.e. evaluation distribution is $\cDsparse{nnz}$ for varying \emph{nnz}. Evaluated after 16 in-context examples.}
\label{fig:linear_nnz}
\begin{subfigure}{.45\textwidth}
    \includegraphics[width=\textwidth]{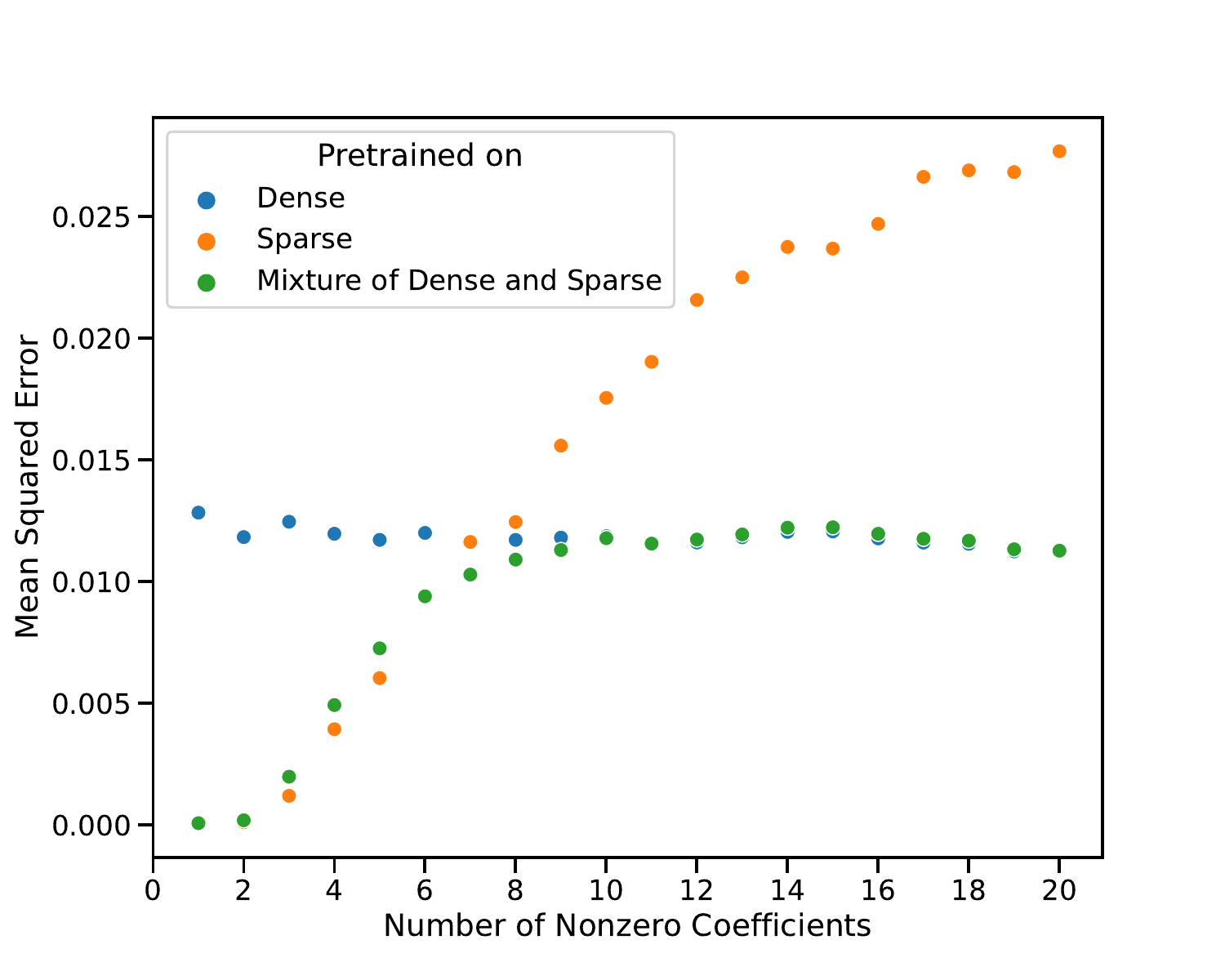}
    \caption{Loss of the models compared. The model pretrained on a mixture of data performs nearly as well as the best of the models pretrained only on one function class or the other.}
    \label{fig:linear_nnz_compare}
\end{subfigure}
\hspace{1em}
\begin{subfigure}{.45\textwidth}
    \includegraphics[width=\textwidth]{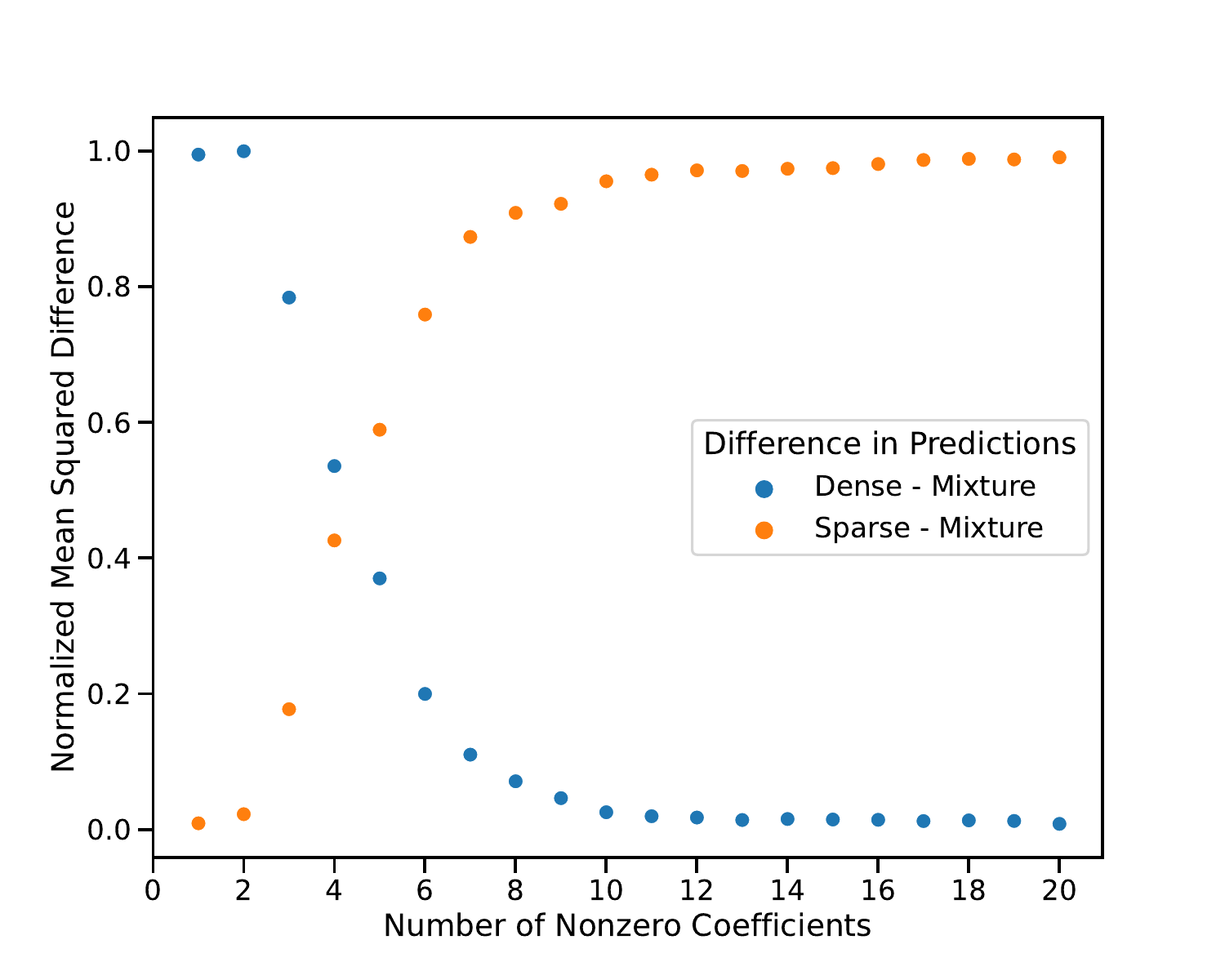}
    \caption{The mean squared difference (MSD) between the predictions from model pretrained on a mixture versus sparse- or dense-only. Normalized by the MSD between the predictions of the sparse and dense models.}
    \label{fig:linear_nnz_pred_diff}
\end{subfigure}
\end{figure}

Returning to the experiment in Figure~\ref{fig:linear_linear_compare} and plotting the error as a function of the number of non-zero coefficients over the entire range of possibilities shows that the model pretrained on the mixture with $w=.5$, $\cDF = 0.5 \cdot \cDdense + 0.5 \cdot \cDsparse{2}$,
performs as well as the models pretrained on the mixture components, ie $w=0$ and $w=1$, throughout (see Figure~\ref{fig:linear_nnz_compare}). This suggests that the model is capable of performing model selection to choose whether to make predictions using knowledge solely from one base function class in the pretraining mixture or the other. Indeed, Figure~\ref{fig:linear_nnz_pred_diff} shows that when the examples provided in-context come from very sparse or very dense functions, the predictions are nearly identical to those made by the models pretrained on either only sparse or only dense data, respectively. In between however, when the number of non-zero coefficients is $\approx 4$, the mixture predictions deviate from both those of the purely dense or purely sparse pretrained transformer. This suggests the model pretrained on the mixture is not simply selecting a single function class for the predictions, but predicting something in between.

\section{Limits of Model Selection Capabilities}
\label{sec:sine}
We examine the ICL generalization capabilities of models along two axes. First, we motivate exploring and test ICL performance on functions the model has both never seen in training and also could plausibly predict: convex combinations of functions drawn from the pretraining function classes. Second, we evaluate ICL performance on functions which are extreme versions of functions seen in pretraining (ie., sinusoids with much higher or lower frequencies than those typically seen in pretraining). In both cases, we find little evidence of out-of-distribution generalization. When the function is significantly far from those seen during pretraining, the predictions are erratic. However, when the function is sufficiently close to the pretraining data, the model approximates it well with predictions from the function classes on which it was pretrained.

Figure~\ref{fig:linear_nnz_compare} shows that the transformer's predictions at moderate sparsity levels (\emph{nnz} = 3 to 7) are not similar to any of the predictions from either of the function classes provided at pretraining, but rather, something in between the two. Hence, one may hypothesize that the model has some inductive bias that allows it to combine pretrained function classes in nontrivial ways. For instance, one may suspect that the model can produce predictions from the combination of the functions that it saw during pretraining. To test this hypothesis in a context with clearly disjoint function classes, we study the ability to perform ICL on linear functions, sinusoids, and convex combinations of the two. We focus on the one dimensional case to make evaluating and visualizing nonlinear function classes straightforward.

\begin{figure}[thb]
\centering     %
\caption{Comparing predictions from models pretrained on different data sources after providing 3 sets of in-context examples (shown in red). Predictions are made by sweeping over $\x$ values passed in as the last element of the sequence after the in-context examples.\\}
\label{fig:one_dim}
\begin{subfigure}[t]{.3\textwidth}
\vskip 0pt
    \includegraphics[width=\textwidth]{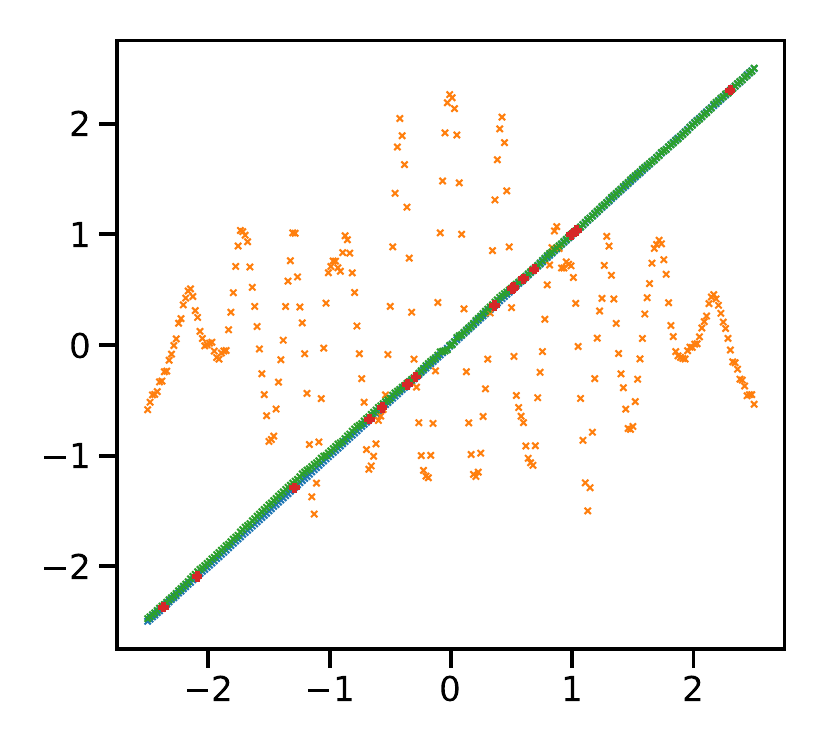}
    \caption{The models pretrained on linear or both linear and sinusoids make good linear predictions when provided examples from a linear model.}
    \label{fig:one_dim_lin}
\end{subfigure}
\hspace{1em}
\begin{subfigure}[t]{.3\textwidth}
\vskip 0pt
    \includegraphics[width=\textwidth]{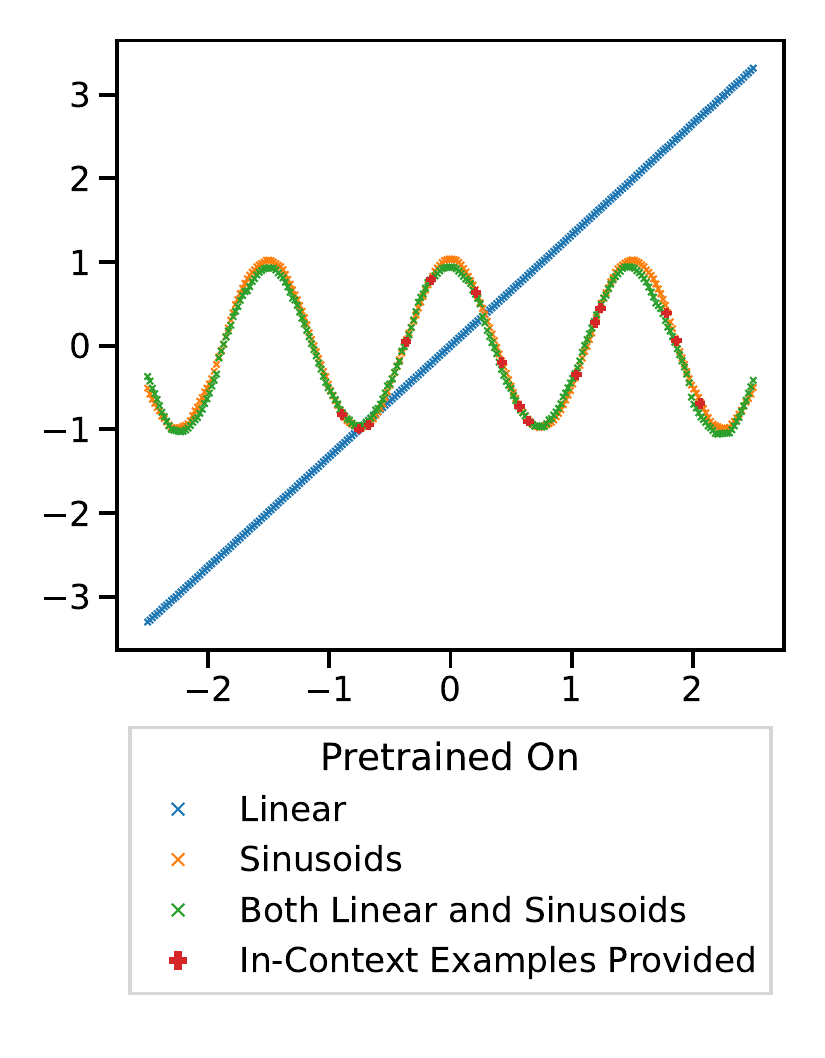}
\caption{The models pretrained on sinusoids or both linear and sinusoids make good sinusoidal predictions when provided examples from a cosine.\\}
    \label{fig:one_dim_sin}
\end{subfigure}
\hspace{1em}
\begin{subfigure}[t]{.3\textwidth}
\vskip 0pt
    \includegraphics[width=\textwidth]{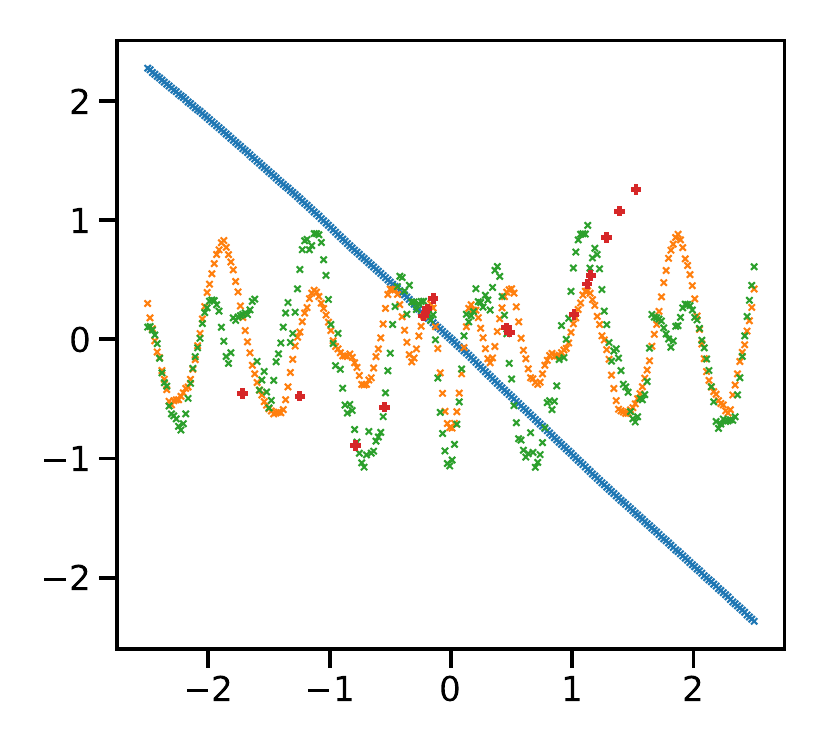}
    \caption{None of the models predict well when provided examples from a convex combination of a linear and cosine functions (although the linear-only model approximates the line of best fit).}
    \label{fig:one_dim_combine}
\end{subfigure}
\end{figure}

Figure~\ref{fig:one_dim} shows that while a model pretrained on a mixture of linear functions and sinusoids (i.e. $\cDF = 0.5 \cdot \cDdense + 0.5 \cdot \cDsine$) is able to make good predictions for either of these functions separately, it is unable to fit a function that is a convex combination of the two\footnote{Note here convex combination refers to adding the weighted y-values of the functions. This is distinct from the weighted mixture distribution used in pretraining which samples either a linear function or sine function for the $f$ used in each prompt sequence.}. This suggests that the interpolation phenomenon shown in Figure~\ref{fig:linear_nnz_pred_diff} for linear functions is not a generalizable inductive bias for in-context learning in transformers. However, it continues to support the narrower hypothesis that when the in-context examples are close to a function class learned in pretraining, the model is capable of selecting the best function class to use for predictions.

\begin{figure}[thb]
\centering     %
\caption{Predictions from transformer models pretrained on linear functions (blue), sinusoids (orange), or both (green) when provided the red examples in-context, coming from a combination of a linear function and sinusoid, with the relative weights noted in the titles of each plot.\\}
\label{fig:superpositions}
\begin{subfigure}[t]{.3\textwidth}
\vskip 0pt
    \includegraphics[width=\textwidth]{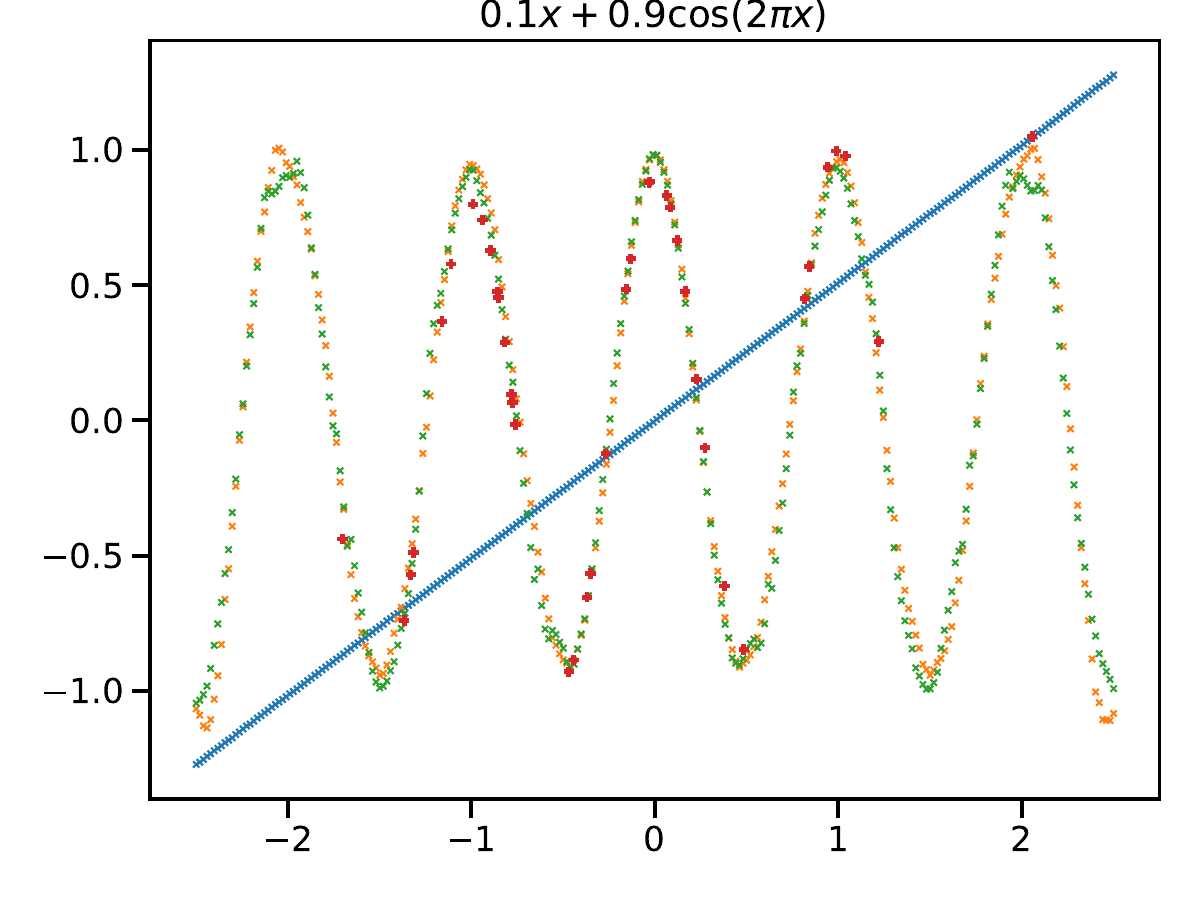}
\end{subfigure}
\hspace{1em}
\begin{subfigure}[t]{.3\textwidth}
\vskip 0pta
    \includegraphics[width=\textwidth]{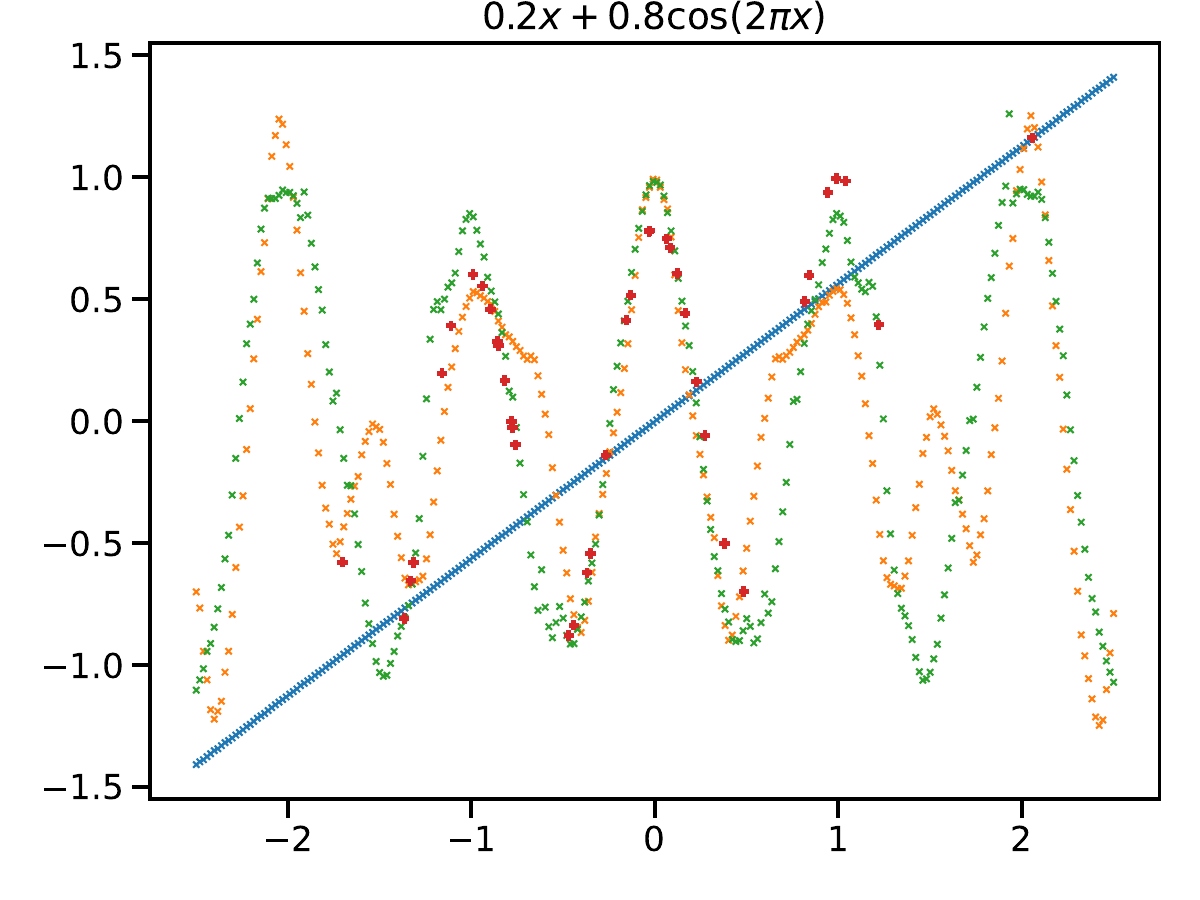}
\end{subfigure}
\hspace{1em}
\begin{subfigure}[t]{.3\textwidth}
\vskip 0pt
    \includegraphics[width=\textwidth]{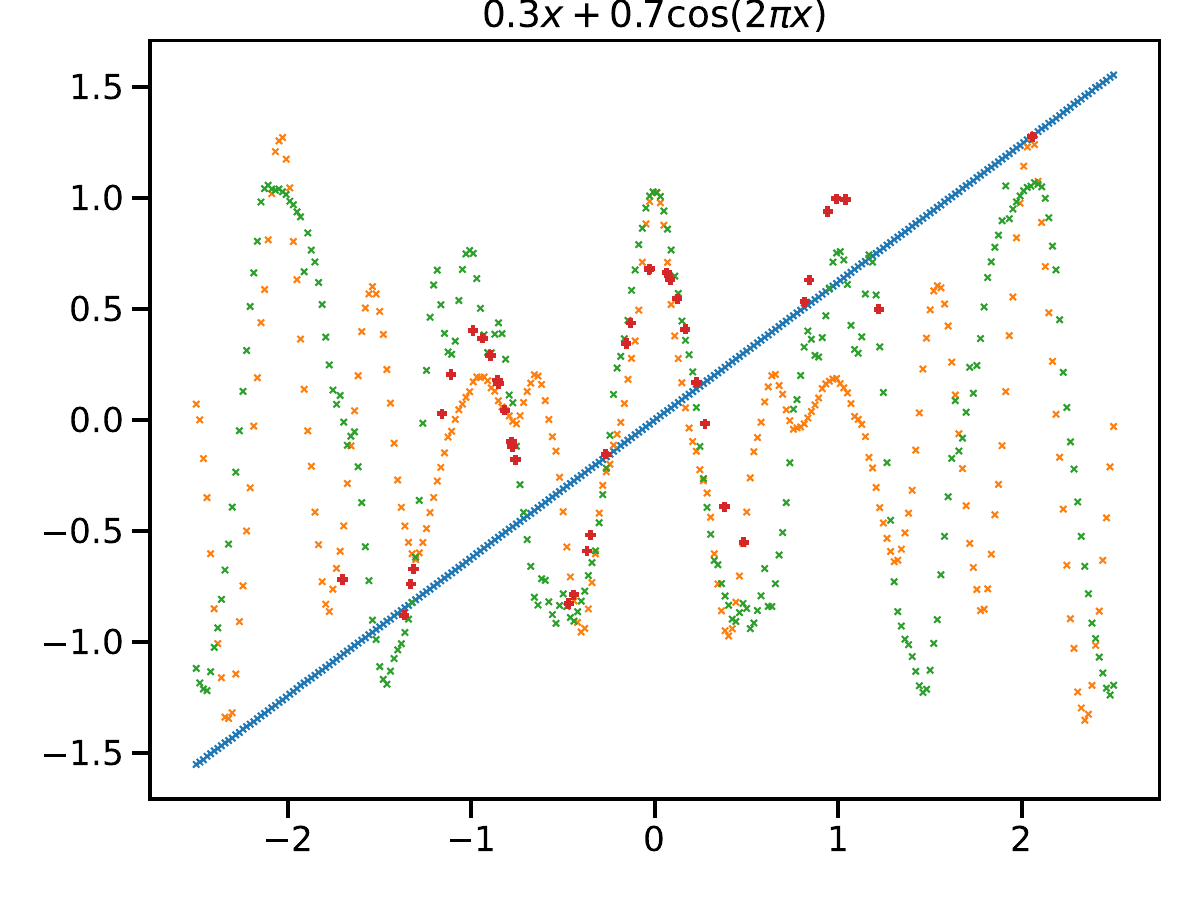}
\end{subfigure}
\begin{subfigure}[t]{.3\textwidth}
\vskip 0pt
    \includegraphics[width=\textwidth]{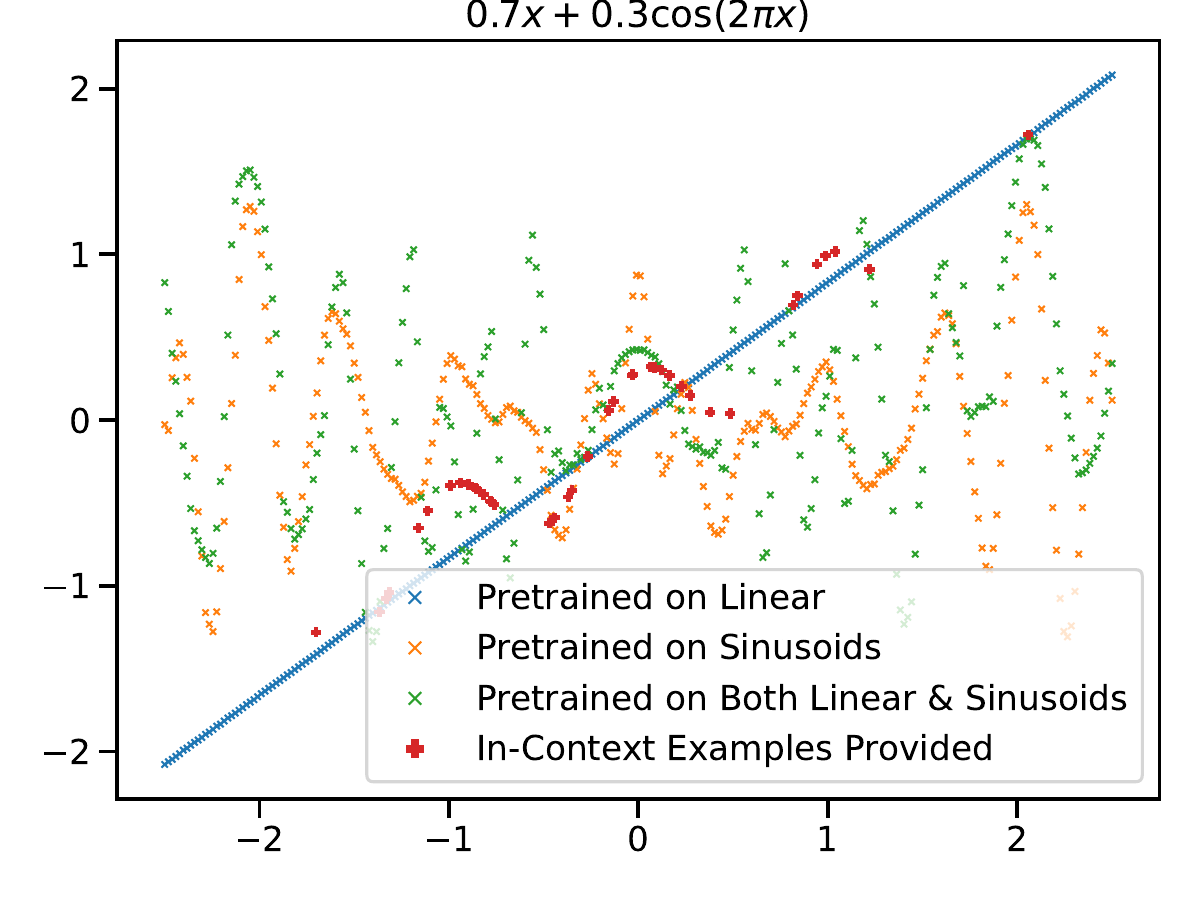}
\end{subfigure}
\hspace{1em}
\begin{subfigure}[t]{.3\textwidth}
\vskip 0pt
    \includegraphics[width=\textwidth]{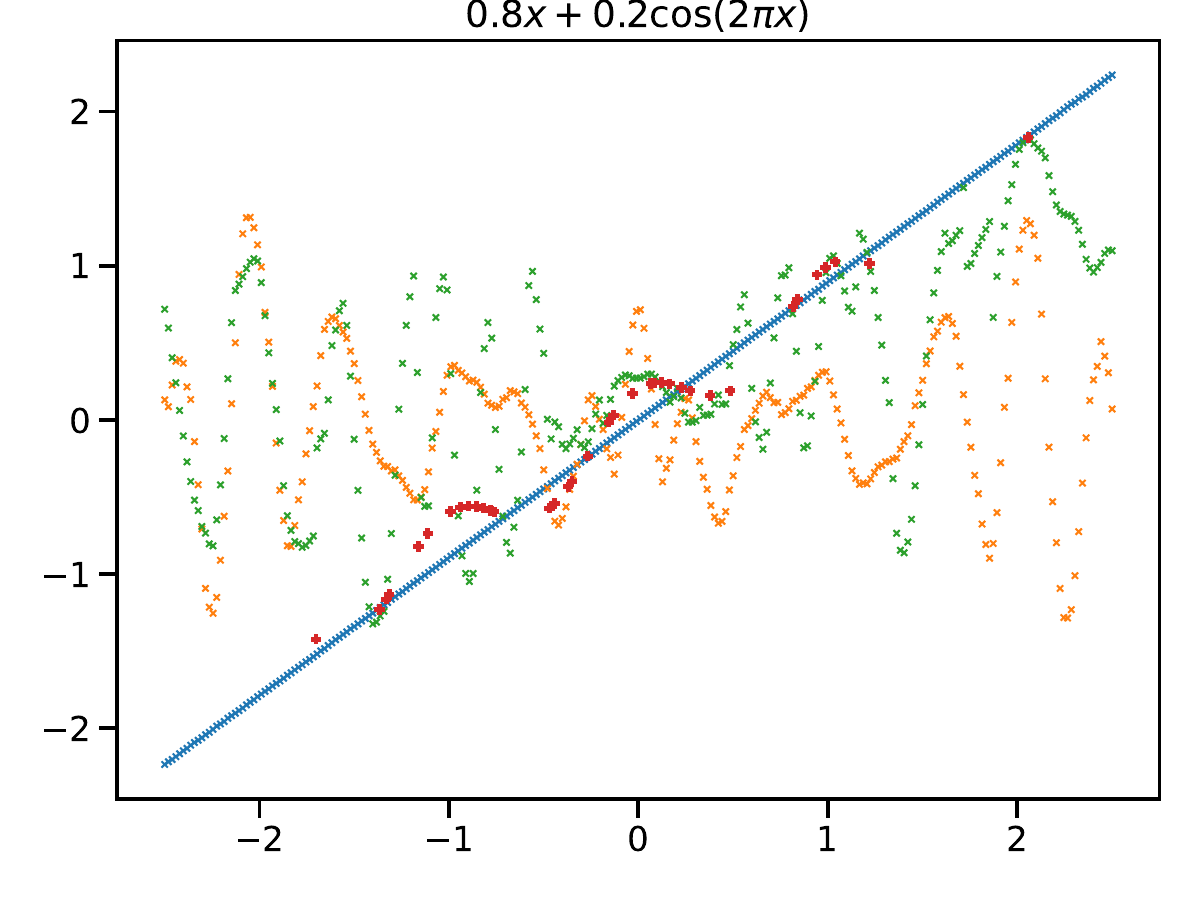}
\end{subfigure}
\hspace{1em}
\begin{subfigure}[t]{.3\textwidth}
\vskip 0pt
    \includegraphics[width=\textwidth]{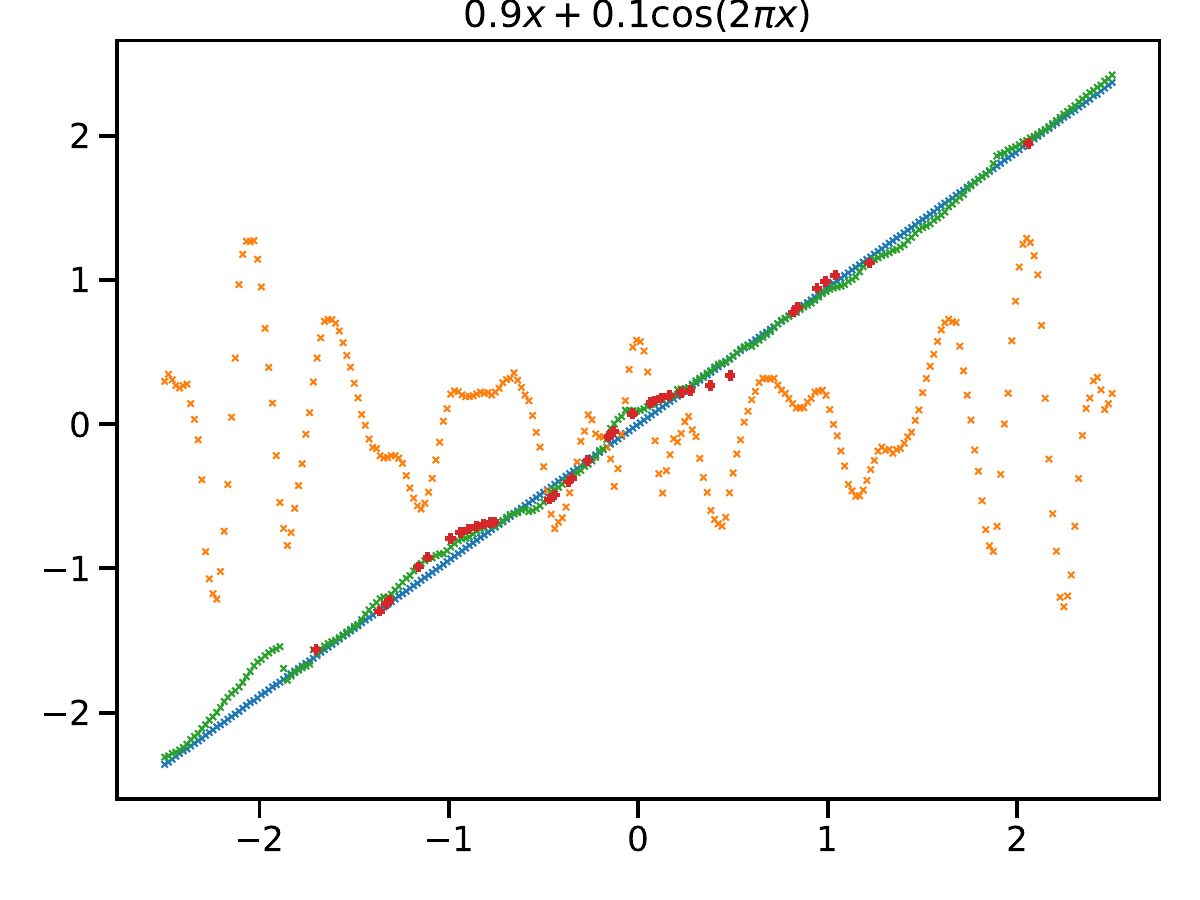}
\end{subfigure}
\end{figure}

Figure~\ref{fig:one_dim_combine} showed a specific convex combination of a linear function and a sinusoid. In Figure~\ref{fig:superpositions}, we sweep over the relative weights of the linear function and sine wave in the convex combination. Here, we observe that when the combined function is predominantly from one function class or the other -- i.e. well-approximated by the function classes learned during pretraining -- the in-context predictions are reasonable. However, when both functions contribute significantly to the convex combination, the model makes erratic predictions not well-justified by the in-context examples. This shows that the model selection capabilities of the model are limited by proximity to the pretraining data, and suggests that broad coverage of function space is critical for generalized in-context learning capabilities.

\begin{figure}[thb]
\centering     %
\caption{Predictions from transformer models pretrained on sinusoids (orange), or both sinusoids and linear functions (green) when provided the red examples in-context, coming from sinusoids of increasing frequency (noted in the plot title).\\}
\label{fig:frequencies}
\begin{subfigure}[t]{.3\textwidth}
\vskip 0pt
    \includegraphics[width=\textwidth]{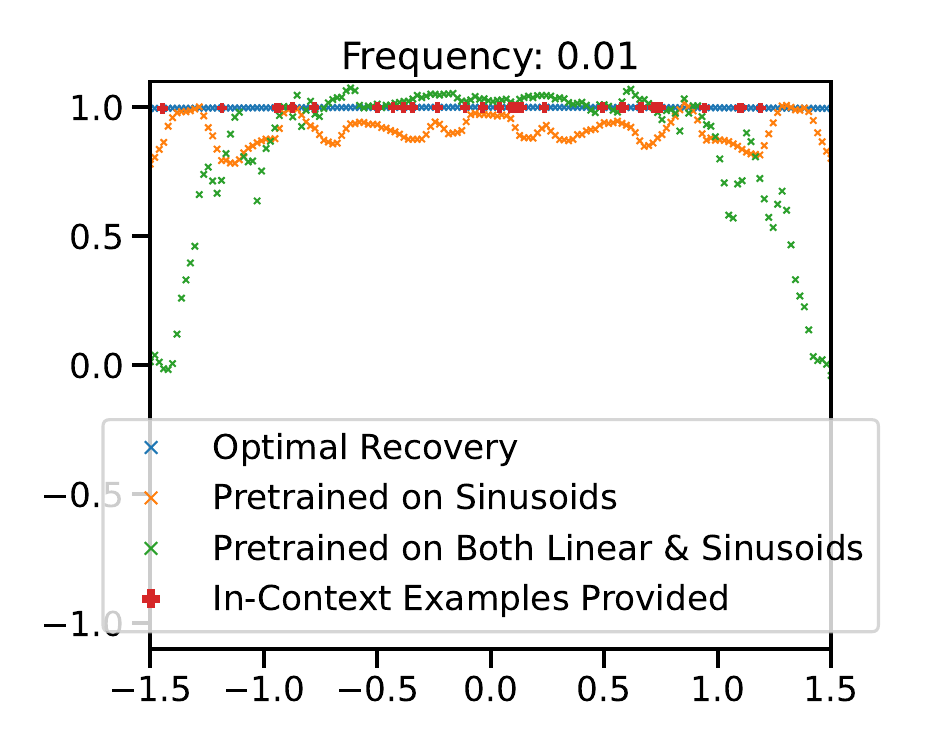}
\end{subfigure}
\hspace{1em}
\begin{subfigure}[t]{.3\textwidth}
\vskip 0pt
    \includegraphics[width=\textwidth]{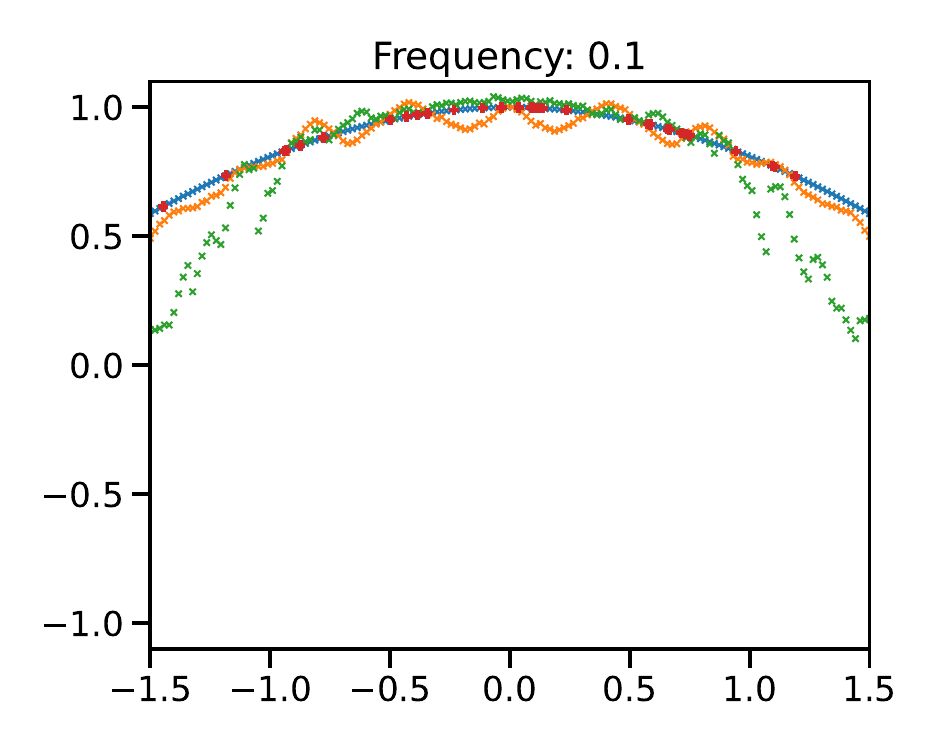}
\end{subfigure}
\hspace{1em}
\begin{subfigure}[t]{.3\textwidth}
\vskip 0pt
    \includegraphics[width=\textwidth]{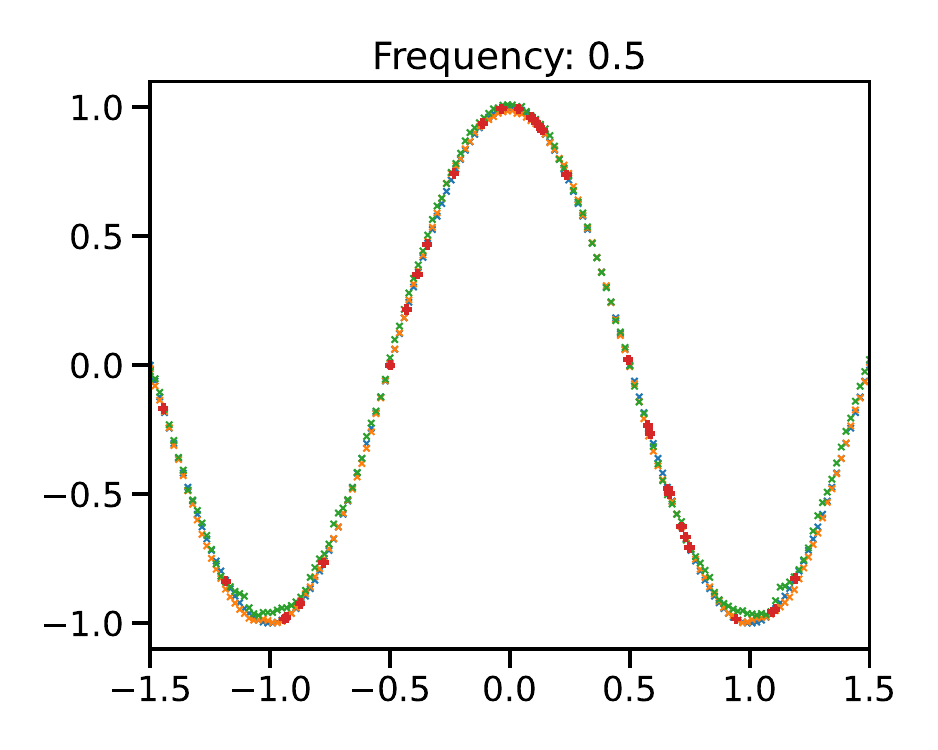}
\end{subfigure}
\begin{subfigure}[t]{.3\textwidth}
\vskip 0pt
    \includegraphics[width=\textwidth]{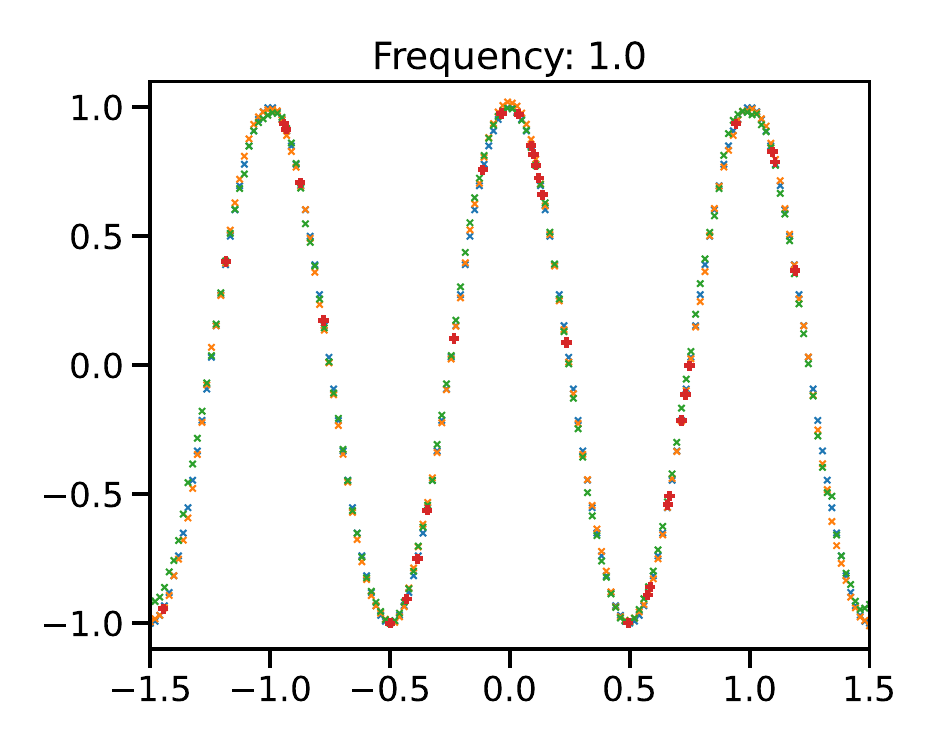}
\end{subfigure}
\hspace{1em}
\begin{subfigure}[t]{.3\textwidth}
\vskip 0pt
    \includegraphics[width=\textwidth]{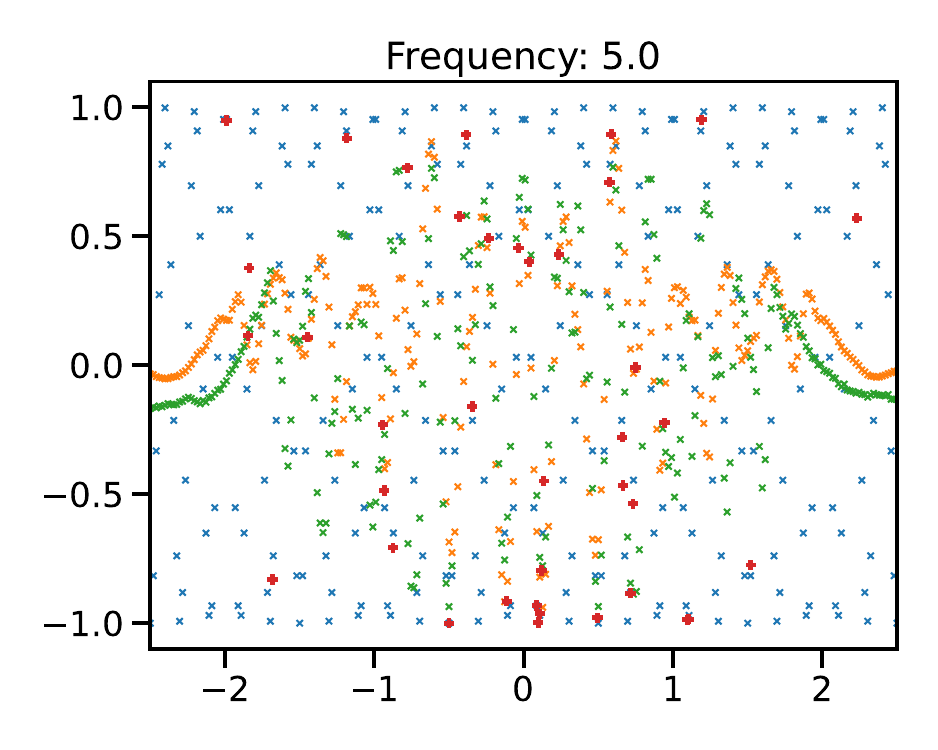}
\end{subfigure}
\end{figure}

The previous convex combinations were specifically constructed so that the model had never seen similar functions in pretraining. Shifting to the scenario where sections of the function class space are increasingly rare in the pretraining data, we find that the model generalization starts strong and then degrades dramatically as the tasks become so rare as to be out-of-distribution. Specifically, we trained models on a mixture of sinusoids with frequencies drawn from a $\textsc{Gamma}(6, 1/6)$ distribution. This distribution ensures that the average frequency is $1$. Frequencies below $0.01$ or above $5$ are extremely rare: the probability of a sampling a frequency outside this range is less than $1 \times 10^{-7}$, i.e. we expect to see approximately 100 examples out of the 1 billion used in pretraining. Figure~\ref{fig:frequencies} shows the predictions from all of the models on a variety of frequencies inside and outside this range.

\section{Conclusion}
We have empirically explored the role of the pretraining data composition on the ability of pretrained transformers to in-context learn function classes both inside and outside the support of their pretraining data distribution. We have empirically shown that for task families or function classes well-represented in the pretraining mixture, cost of selecting the appropriate function class to use for in-context learning is nearly zero. We next explored generalizability on two scenarios: (1) We found that the pretrained transformers struggle to predict on convex combinations of functions drawn from pre-training function classes, and (2) We observed that transformers can generalize effectively on rarer sections of the function-class space and still break down as the tasks become out-of-distribution. 

An important question is understanding how the observations we make here carry over to tokenized models and to questions represented in natural language. We attempted an experiment to train a tokenized model for the one-dimensional examples presented in Section~\ref{sec:sine} by binning the scalar values into buckets, and treating the bucket indices as tokens for the input to a transformer-based language model. We trained this model for 5M epochs with a cross-entropy loss as typically used in language models, but were unable to significantly decrease the loss. Understanding the challenges to training such a model and evaluating whether this framing has different model selection or out-of-distribution generalization properties is important future work. In the natural language setting, the intuitive notions of how to appropriately define task families (in our case function classes), precise pretraining mixtures, and convex combinations are all less clear. We believe bridging the gap between the notions presented here and in language modeling may help to improve our understanding of the power of ICL and how to effectively enable it.

\subsection*{References}
\bibliographystyle{abbrvnat}
\bibliography{bib}

\begin{thebibliography}{7}
\providecommand{\natexlab}[1]{#1}
\providecommand{\url}[1]{\texttt{#1}}
\expandafter\ifx\csname urlstyle\endcsname\relax
  \providecommand{\doi}[1]{doi: #1}\else
  \providecommand{\doi}{doi: \begingroup \urlstyle{rm}\Url}\fi

\bibitem[Aky{\"u}rek et~al.(2022)Aky{\"u}rek, Schuurmans, Andreas, Ma, and
  Zhou]{akyurek2022learning}
E.~Aky{\"u}rek, D.~Schuurmans, J.~Andreas, T.~Ma, and D.~Zhou.
\newblock What learning algorithm is in-context learning? investigations with
  linear models.
\newblock In \emph{The Eleventh International Conference on Learning
  Representations}, 2022.

\bibitem[Bai et~al.(2023)Bai, Chen, Wang, Xiong, and Mei]{bai2023transformers}
Y.~Bai, F.~Chen, H.~Wang, C.~Xiong, and S.~Mei.
\newblock Transformers as statisticians: Provable in-context learning with
  in-context algorithm selection, 2023.

\bibitem[Brown et~al.(2020)Brown, Mann, Ryder, Subbiah, Kaplan, Dhariwal,
  Neelakantan, Shyam, Sastry, Askell, Agarwal, Herbert-Voss, Krueger, Henighan,
  Child, Ramesh, Ziegler, Wu, Winter, Hesse, Chen, Sigler, Litwin, Gray, Chess,
  Clark, Berner, McCandlish, Radford, Sutskever, and
  Amodei]{NEURIPS2020_1457c0d6}
T.~Brown, B.~Mann, N.~Ryder, M.~Subbiah, J.~D. Kaplan, P.~Dhariwal,
  A.~Neelakantan, P.~Shyam, G.~Sastry, A.~Askell, S.~Agarwal, A.~Herbert-Voss,
  G.~Krueger, T.~Henighan, R.~Child, A.~Ramesh, D.~Ziegler, J.~Wu, C.~Winter,
  C.~Hesse, M.~Chen, E.~Sigler, M.~Litwin, S.~Gray, B.~Chess, J.~Clark,
  C.~Berner, S.~McCandlish, A.~Radford, I.~Sutskever, and D.~Amodei.
\newblock Language models are few-shot learners.
\newblock In H.~Larochelle, M.~Ranzato, R.~Hadsell, M.~Balcan, and H.~Lin,
  editors, \emph{Advances in Neural Information Processing Systems}, volume~33,
  pages 1877--1901. Curran Associates, Inc., 2020.
\newblock URL
  \url{https://proceedings.neurips.cc/paper_files/paper/2020/file/1457c0d6bfcb4967418bfb8ac142f64a-Paper.pdf}.

\bibitem[Garg et~al.(2022)Garg, Tsipras, Liang, and Valiant]{garg2022can}
S.~Garg, D.~Tsipras, P.~S. Liang, and G.~Valiant.
\newblock What can transformers learn in-context? a case study of simple
  function classes.
\newblock \emph{Advances in Neural Information Processing Systems},
  35:\penalty0 30583--30598, 2022.

\bibitem[Li et~al.(2023)Li, Emrullah~Ildiz, Papailiopoulos, and
  Oymak]{li2023transformers}
Y.~Li, M.~Emrullah~Ildiz, D.~Papailiopoulos, and S.~Oymak.
\newblock Transformers as algorithms: Generalization and stability in
  in-context learning.
\newblock \emph{arXiv e-prints}, pages arXiv--2301, 2023.

\bibitem[Raventós et~al.(2023)Raventós, Paul, Chen, and
  Ganguli]{raventos2023pretraining}
A.~Raventós, M.~Paul, F.~Chen, and S.~Ganguli.
\newblock Pretraining task diversity and the emergence of non-bayesian
  in-context learning for regression, 2023.

\bibitem[Vaswani et~al.(2017)Vaswani, Shazeer, Parmar, Uszkoreit, Jones, Gomez,
  Kaiser, and Polosukhin]{DBLP:journals/corr/VaswaniSPUJGKP17}
A.~Vaswani, N.~Shazeer, N.~Parmar, J.~Uszkoreit, L.~Jones, A.~N. Gomez,
  L.~Kaiser, and I.~Polosukhin.
\newblock Attention is all you need.
\newblock \emph{CoRR}, abs/1706.03762, 2017.
\newblock URL \url{http://arxiv.org/abs/1706.03762}.

\end{thebibliography}
\appendix
\section{Architecture Training and Data Generation}

\subsection{Architecture and Training}
\label{app:architecture}
Following \citet{garg2022can}'s approach to convert a sequence of $n$ alternating $d$-dimensional $\x$ values and $1$-dimensional $f(\x)$ values into a single embedded sequence for the transformer, we make $f(\x)$ $d$-dimensional by padding it with 0s, producing a sequence of $2n$ "tokens" of $\x_i$ and $f(\x_i)$, each of which is a $d$-dimensional real-valued vector. We insert a (learnable) linear layer to project each input into the transformer's model dimension. Correspondingly, we insert another (learnable) linear layer to reverse the embedding and produce a $d$-dimensional output representing $\x$ or $f(\x)$, depending on the sequence position. The loss at training time uses the next-token prediction squared loss computed over alternating positions: that is, the loss is only computed over the $\x_i$ positions and ignores the $\x_{i+1}$ value predicted by the transformer structure for each $f(\x_{i})$ in the input. As in prior work, the model is pretrained on simulated data for different data-generating setups. We do not fine-tune a pretrained language model and do not train on actual text. Concretely, each training batch consists of $k$ prompts. Each prompt is a sequence of $\x_i, f(\x_i)$ pairs, with the $\x$'s sampled from $\cD_{\cX}$ (i.e. $\mathcal{N}(0, I_d)$) and a single $f \sim \cD_{\cF}$ per prompt. We use a fixed $w$ per pretraining run. 

We trained a decoder-only Transformer \citep{DBLP:journals/corr/VaswaniSPUJGKP17} model of GPT-2 scale implemented in the Jax-based machine learning framework, Pax\footnote{\url{https://github.com/google/paxml}} with 12 layers, 8 attention heads, and a 256-dimensional embedding space (9.5M parameters) as our base configuration \citep{garg2022can}. We too set the dimensions of $x$ as 20 (except when specified otherwise) and used standard cosine positional embeddings in our model. We trained 1 million steps with a training batch size of 1024. We used the Adam optimizer with standard values set to $\beta_1=0.9$, $\beta_2=0.999$, $\epsilon=10^{-9}$ and no weight-decay. We used a linear ramp-up schedule followed by an inverse-square root learning rate decay. We empirically tuned to find a learning rate of 1 with 5,000 warm-up steps to be effective. We present our results without noise in the data generation although find similar results adding label noise.

\subsection{Data Generation and Function Classes}
\label{app:data_gen}
Throughout our paper we always use $\mathcal{N}(0, \I_d)$ for the covariate distribution (so each $\x_i \sim \mathcal{N}(0, \I_d)$ and consider models in dimension $d=20$ (for linear and high-dimensional functions) or $d=1$ (for the 1-dimensional sinusoidal examples). For the function classes considered in the text we use:
\begin{itemize}
    \item For the class of dense linear functions $\cDdense$ we generate a random $\bbeta \sim \cN(0, \I_d)$ and define $\cDdense = \{ f(\x_i) : f(\x_i) = \bbeta^\top \x_i/\sqrt{d} \}$.
    \item For $\cDsparse{nnz}$ we consider sparse linear models with the number of non-zero elements set to a sparsity level denoted by \emph{s} or \emph{nnz}. To generate the underlying parameter vector we randomly sample $s$ coordinates uniformly without replacement from the $d=20$ dimensional support and in each non-zero coordinate generate a random coefficient with distribution $\cN(0,1)$ to create $\bbeta$. We then define $\cDsparse{s} = \{ f(\x_i) : f(\x_i) = \bbeta^\top \x_i/\sqrt{s} \}$. 
    \item For $\cDrelu$ we consider two-layer ReLU networks with randomly generated weights. In particular we generate a first layer weight matrix $\W \in \mathbb{R}^{d_h \times d}$ where $d_h = 100$ is the hidden dimension and $\W_{ij} \sim \cN(0, 2)$ and second layer coefficient $\bbeta \sim \cN(0,\I_{d_h})$. We then define the function class as $\cDrelu = \{f(\x_i): f(\x_i) = \bbeta^\top \sigma(\W \x_i)/(\sqrt{d \cdot d_h})\}$, where the activation function $\sigma(\cdot)$ is taken as the standard ReLU unit.
    \item To sample $f \sim \cDsine$, first, we sample $\omega \sim \Gamma(6, 1/6)$, then sample $\beta \sim \cN(0, 1)$, and let $f(x) = \beta \cos( 2 \pi \omega x)$.
\end{itemize}
Note that in each case we define the normalization of each function class such that $\text{Var}(f(\x_i))$ are equalized to ensure the norms of the outputs are comparable across different function classes.

\section{Additional Model Selection Experiments}
\label{app:model_selection}
\begin{figure}[thb]
\centering
\caption{\label{fig:dense_relu_learning_curves}ICL learning curves displaying mean-squared error (MSE) for evaluations on prompts drawn from $\cD_{dense}$ (left) and $\cD_{ReLU}$ (right). Transformers were pretrained on mixtures of $w \cdot \cD_{dense} + (1-w) \cdot \cD_{ReLU}$. Different curves correspond to different $w$.}
\begin{subfigure}{.45\textwidth}
    \includegraphics[scale=0.3]{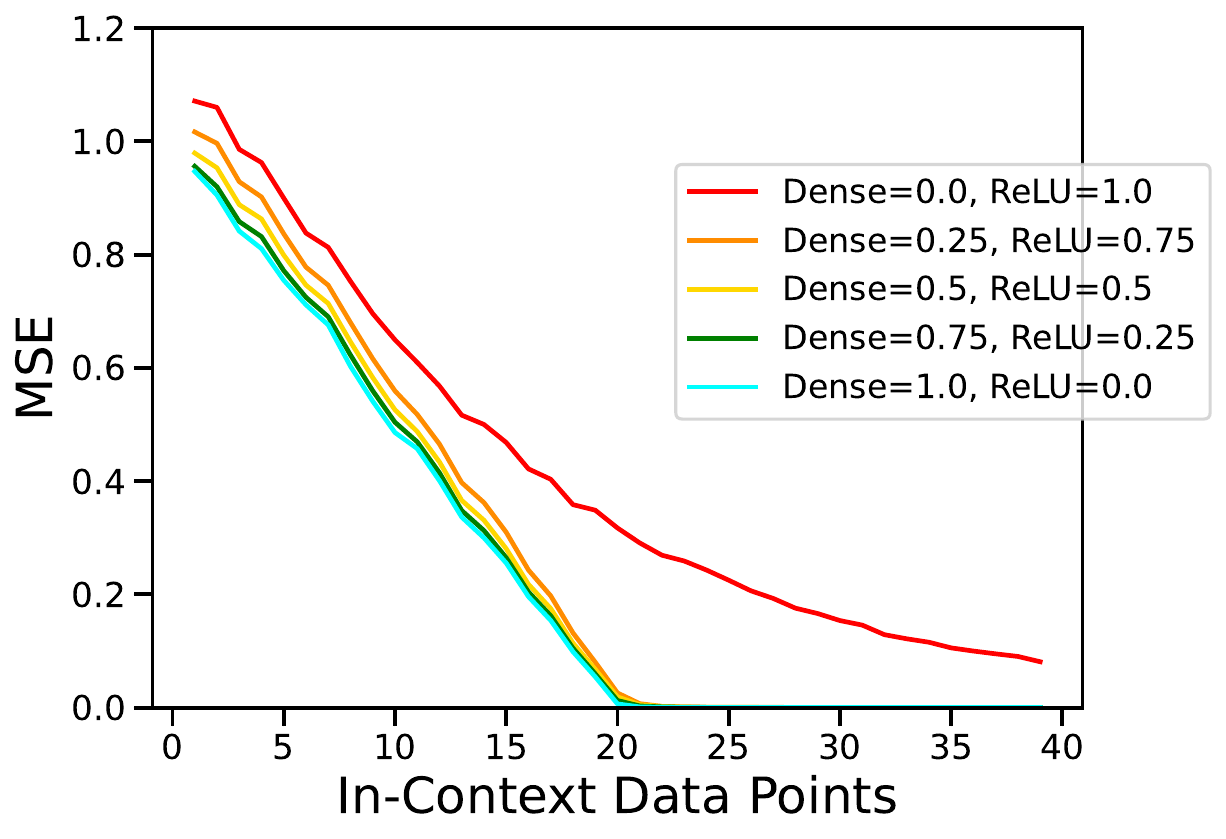}
    \caption{Evaluated on prompts from $\cD_{dense}$}
    \label{subfig:dense_relu_dense}
\end{subfigure}
\begin{subfigure}{.45\textwidth}
    \includegraphics[scale=0.3]{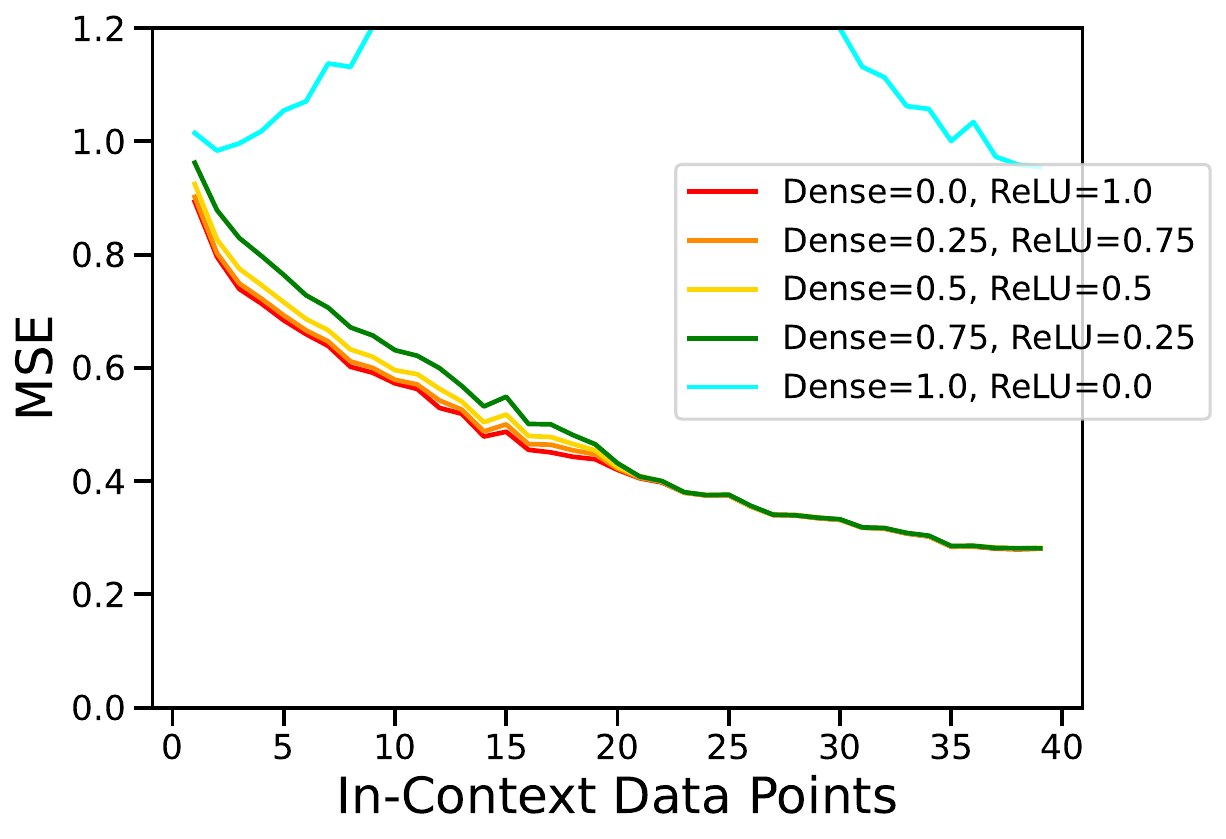}
    \caption{Evaluated on prompts from $\cD_{ReLU}$}
    \label{subfig:dense_relu_relu}
\end{subfigure}
\end{figure}

We produce the ICL learning curves in Figure \ref{fig:dense_relu_learning_curves} for a mixture of dense linear data and data generated from the ReLU function class to see similar phenomenon: in-context model selection cost is small relative to the baseline of only training on that single function class. Figure~\ref{fig:dense_relu_learning_curves} shows this for any pretraining mixture that has non-trivial weight $w$ (or $1-w$) on the prompts it is evaluated. We again observe only small deviations and these decay quickly as ICL sample count increases.
As before, transformer model ICL generalization suffers out-of-distribution, i.e. when evaluating on the function class not seen in pretraining. We do note in Figure~\ref{subfig:dense_relu_dense} the ICL error does improve as the number of in-context data points increases. This is likely because the ReLU function class can locally approximate the linear function class for large numbers of samples.

\begin{figure}[thb]
\centering
\caption{\label{fig:multisize_dense_sparse}ICL learning curves  displaying mean-squared error (MSE) for evaluations on prompts drawn from $\cD_{dense}$ (left) and $\cD_{sparse, nnz=2}$ (right). Transformers of varying sizes pretrained on mixtures with $w=.5$ in $w \cdot \cD_{dense} + (1-w) \cdot \cD_{sparse, nnz=2}$. }
\begin{subfigure}{.45\textwidth}
    \includegraphics[scale=0.3]{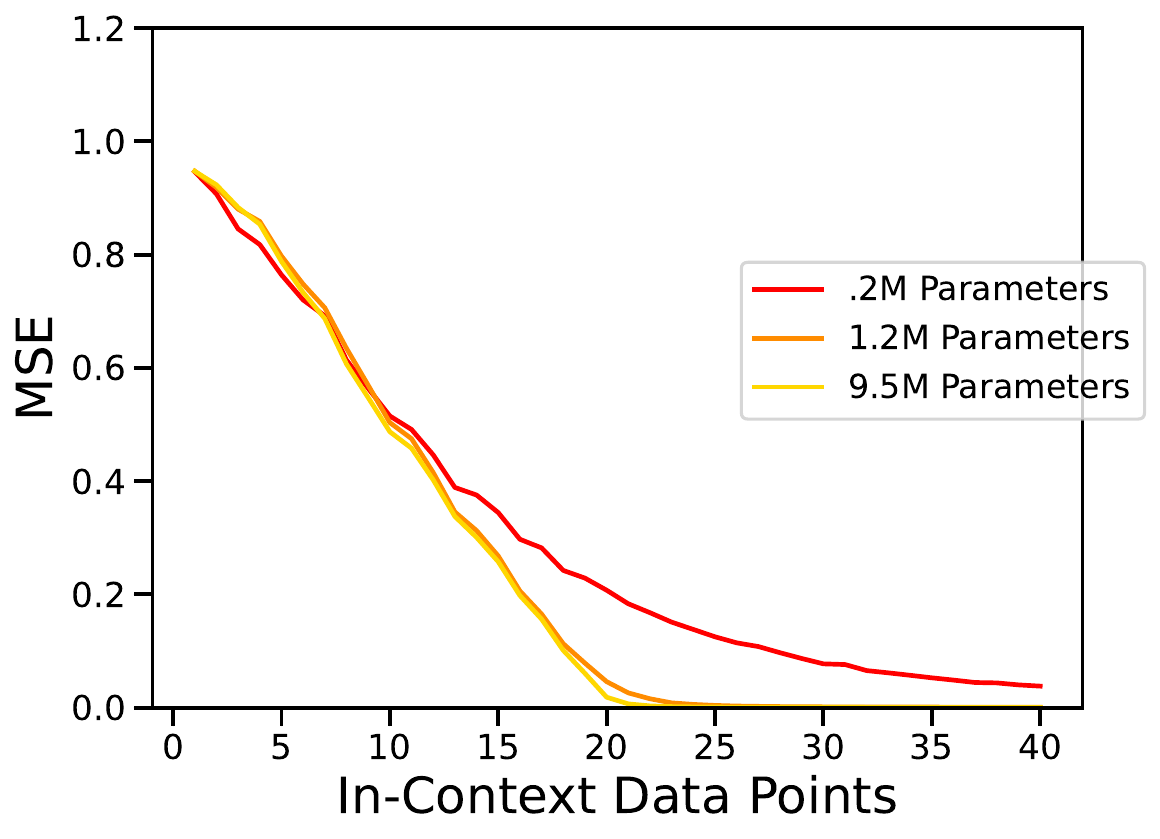}
    \caption{Evaluated on prompts from $\cD_{dense}$}
    \label{subfig:multisize_dense_sparse_a}
\end{subfigure}
\begin{subfigure}{.45\textwidth}
    \includegraphics[scale=0.3]{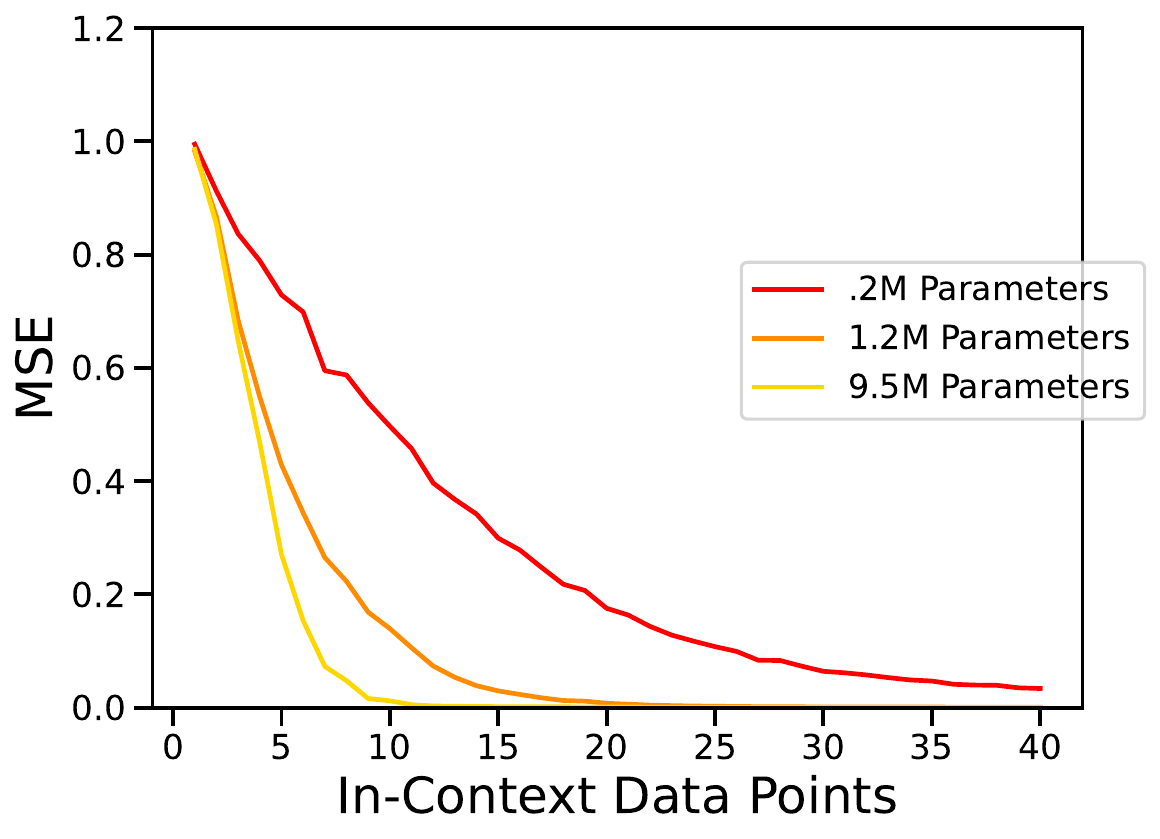}
    \caption{Evaluated on prompts from $\cD_{sparse, nnz=2}$}
    \label{subfig:multisize_dense_sparse_b}
\end{subfigure}
\end{figure}

\begin{figure}[thb]
\centering
\caption{\label{fig:multisize_dense_relu}ICL learning curves displaying mean-squared error (MSE) for evaluations on prompts drawn from $\cD_{dense}$ (left) and $\cD_{ReLU}$ (right). Transformers of varying sizes pretrained on mixtures with $w=.5$ in $w \cdot \cD_{dense} + (1-w) \cdot \cD_{ReLU}$.}
\begin{subfigure}{.45\textwidth}
    \includegraphics[scale=0.3]{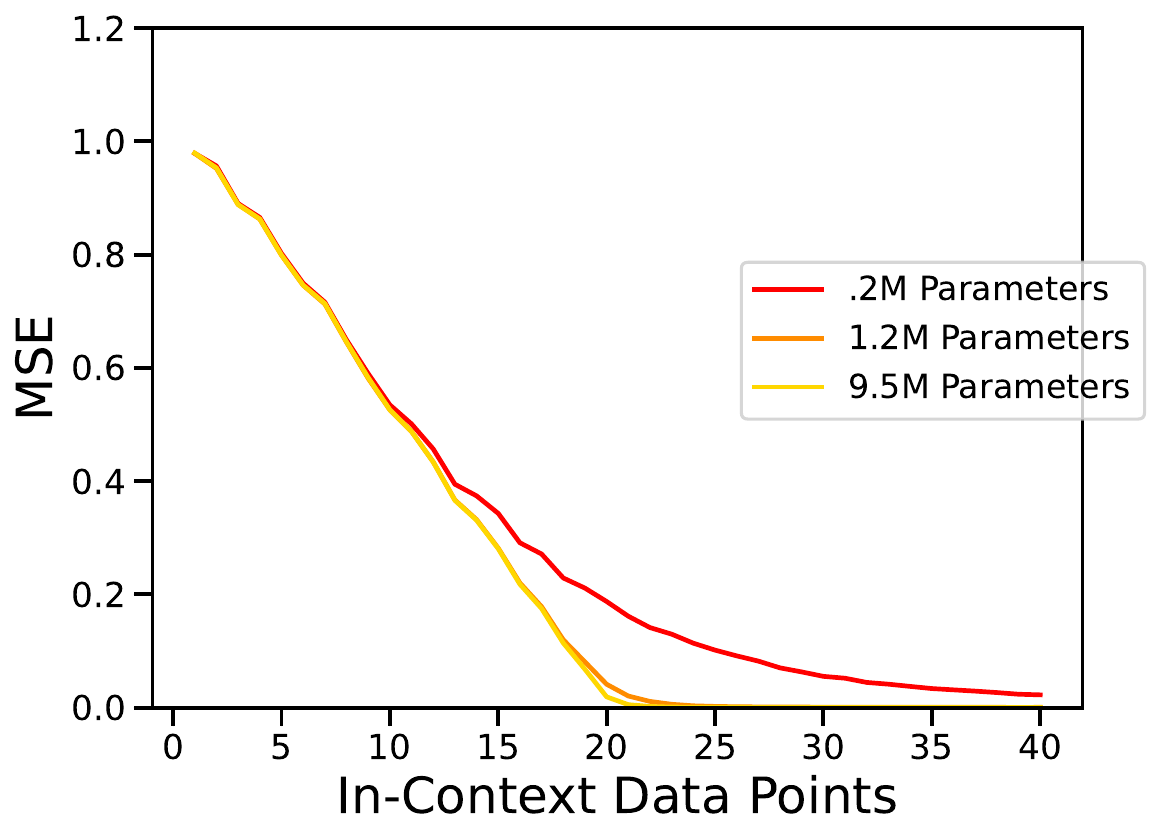}
    \caption{Evaluated on prompts from $\cD_{dense}$}
    \label{subfig:multisize_dense_relu_a}
\end{subfigure}
\begin{subfigure}{.45\textwidth}
    \includegraphics[scale=0.3]{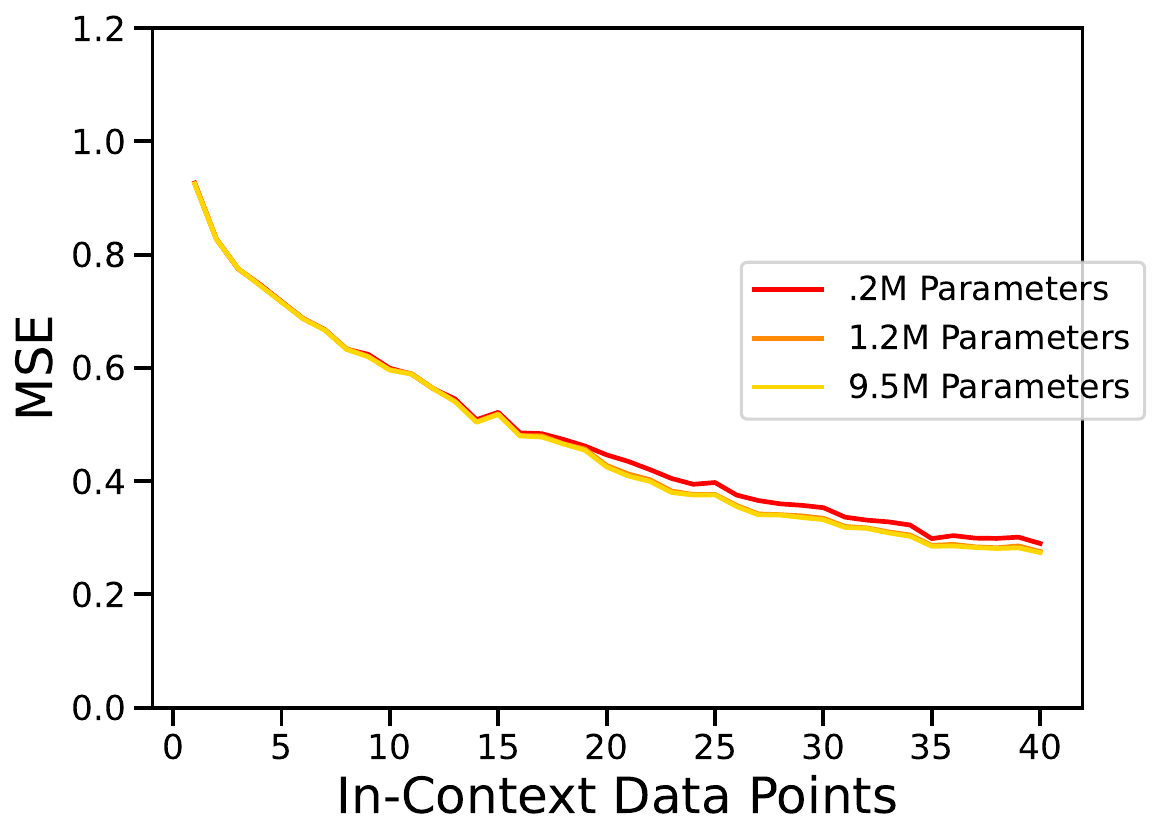}
    \caption{Evaluated on prompts from $\cD_{ReLU}$}
    \label{subfig:multisize_dense_relu_b}
\end{subfigure}
\end{figure}

Exploring the effect of model size, we observe the general trend that the ability to perform model selection increases as models get larger. Additionally, we observe the behavior is asymmetric. Figure~\ref{subfig:multisize_dense_sparse_a} and Figure~\ref{subfig:multisize_dense_relu_a} show that the smallest .2M parameter model performs relatively poorly while the 1.2M and 9.5M deliver similar performance on this problem. However, Figure~\ref{subfig:multisize_dense_sparse_b} exhibits both a larger variation in performance across sizes and significantly worse performances by .2M and 1.2M vs 9.5M on prompts from $\cD_{sparse, nnz=2}$ compared to prompts from $\cD_{dense}$. Meanwhile, Figure~\ref{subfig:multisize_dense_relu_b} shows no difference in performance as a function of model size.

We used the same model sizes and architectures as \citet{garg2022can}'s Section A.1 with models Tiny, Small, and Standard.

\end{document}